\documentclass[journal]{IEEEtran}
\usepackage{amssymb,amsmath,epsfig,array, graphicx, tipa, color, verbatim, caption, subcaption}
\usepackage{mathrsfs}
\usepackage{tabularx}
\usepackage{multirow}

\usepackage[utf8]{inputenc}
\usepackage[vietnam, english]{babel}

\begin{document}
\title{Phonology-Augmented Statistical Framework\\ for Machine Transliteration using Limited Linguistic Resources}

\author{Gia H. Ngo,  Minh Nguyen, Nancy F. Chen
\thanks{Gia H. Ngo is currently with Cornell University, but part of this work was done at the Institute for Infocomm Research, A*STAR. Minh Nguyen is currently with National University of Singapore. Nancy F. Chen is currently with the Institute for Infocomm Research, A*STAR.\newline\hspace*{1em}\textcopyright~ 2018 IEEE. Personal use of this material is permitted. Permission from IEEE must be obtained for all other uses, in any current or future media, including reprinting/republishing this material for advertising or promotional purposes, creating new collective works, for resale or redistribution to servers or lists, or reuse of any copyrighted component of this work in other works.}}

\markboth{IEEE TRANSACTIONS ON AUDIO, SPEECH, AND LANGUAGE PROCESSING}%
{Shell \MakeLowercase{\textit{et al.}}: Bare Demo of IEEEtran.cls for Journals}

\maketitle

\begin{abstract}
Transliteration converts words in a source language (e.g., English) into words in a target language (e.g., Vietnamese).
This conversion considers the phonological structure of the target language, as the transliterated output needs to be pronounceable in the target language.
For example, a word in Vietnamese that begins with a consonant cluster is phonologically invalid and thus would be an incorrect output of a transliteration system.
Most statistical transliteration approaches, albeit being widely adopted, do not explicitly model the target language's phonology, which often results in invalid outputs. 
The problem is compounded by the limited linguistic resources available when converting foreign words to transliterated words in the target language.
In this work, we present a phonology-augmented statistical framework suitable for transliteration, especially when only limited linguistic resources are available.
We propose the concept of pseudo-syllables as structures representing how segments of a foreign word are organized according to the syllables of the target language's phonology.
We performed transliteration experiments on Vietnamese and Cantonese.
We show that the proposed framework outperforms the statistical baseline by up to 44.68\% relative, when there are limited training examples (587 entries).
\end{abstract}


\IEEEpeerreviewmaketitle

\section{Introduction}

\IEEEPARstart{I}{n} every language, new words are constantly being invented or borrowed from foreign languages (e.g. names of people, locations, organizations, and products). 
For example, the city's name ``Manchester'' has become well known by people of languages other than English.
These new words are often named entities that are important in  cross-lingual information retrieval \cite{meng2001generating}\cite{fujii2001japanese}\cite{virga2003}\cite{abduljaleel2003statistical}\cite{oh2006ensemble},
information extraction \cite{marton2014transliteration},
 machine translation \cite{knight1998}\cite{al2002translating}\cite{hermjakob2008name}\cite{durrani2014improving}\cite{durrani2014integrating},
 and often present  out-of-vocabulary challenges to spoken language technologies such as automatic speech recognition \cite{Mansikkaniemi2012},
 spoken keyword search \cite{schone2006low}\cite{zhang2010automatic}\cite{chen2014kws},
 and text-to-speech \cite{Eklund1998}\cite{jung2000english}.
Transliteration is a mechanism for converting a word in a source (foreign) language to a target language, and often adopts approaches from machine translation. 
In machine translation, the objective is to preserve the semantic meaning of the utterance as much as possible while following the syntactic structure in the target language.
In transliteration, the objective is to preserve the original pronunciation of the source word as much as possible while following the phonological structures of the target language.

The amount of training data available for transliteration is often much less than that of machine translation. The amount of training data for machine translation is not limited to the adoption of new vocabulary or concepts from a foreign language while this is true for transliteration. It is therefore challenging for a statistical model to generalize well to implicitly learn the phonological rules of the target language for transliteration tasks. 
The lack of training data often results in non-interpretable outputs by statistical transliteration models \cite{Yoon2006}; these outputs are invalid because speakers of the target language are unable to pronounce these transliterated outputs.
Given the limited training data for transliteration, performance of statistical transliteration approaches has often been suboptimal \cite{knight1998}\cite{al2002machine}.
On the other hand, symbolic transliteration approaches have been shown to produce phonologically-valid outputs with minimal training resources \cite{ngo2014minimal}.
However, symbolic approaches are often limited by the complexity of the predefined rules, and therefore, under-perform with larger datasets, as compared to statistical methods \cite{ngo2014minimal}.

We propose a transliteration framework in which n-gram language modeling is augmented with phonological knowledge.
We propose the concept of pseudo-syllables in statistical models to impose phonological constraints of syllable structure in the target language, yet retain acoustic authenticity of the source language as closely as possible.
Our proposed framework integrates advantages of symbolic approaches on top of statistical transliteration models.
The proposed approach ensures phonologically-valid outputs, while maintaining strengths of statistical models (e.g., language-independence, performance scaling up with training data size). 

This work extends and expands our prior work  \cite{ngo2014minimal} to include detailed formulations, experiments, analyses, and discussions left out in the conference version. 
In particular, empirical validation has been generalized on two language pairs: English-to-Vietnamese and English-to-Cantonese, using the Vietnamese corpora from the IARPA BABEL program\footnote{https://catalog.ldc.upenn.edu/LDC2017S01} that was released for the NIST OpenKWS13 Evaluation \cite{openkws2013} and for the shared tasks at the Named Entity Workshop (NEWS) at ACL 2018\footnote{The English-to-Vietnamese transliteration data in \cite{ngo2014minimal} was released at NEWS 2018 (http://workshop.colips.org/news2018/shared.html)}. The implementation of the proposed model, the symbolic systems and the customized tools for evaluating transliteration error rates are publicly available\footnote{https://github.com/ngohgia/transliteration}.

\section{Background}
\label{sec:background}
\subsection{Phonology}
\label{subsec:phonology}
\noindent We introduce three phonological concepts relevant to our discussion.
\vspace{0.1cm}

\subsubsection{Syllable}
\label{subsubsec:background_syllable}
\noindent A syllable is considered the smallest phonological unit of a word \cite{ladefoged2014} with the following structure \cite{goldsmith2011handbook}\cite{kessler1997syllable}:
\vspace{-0.1cm}
\begin{align}
  \left[O\right]N\left[Cd\right] + \left[T\right]
  \label{syl_struct}
\end{align}
\noindent where the ``[ ]'' specifies an optional unit. 
$O$ denotes the Onset, which is a consonant or a cluster of consonants at the beginning of a syllable.
$N$ denotes the Nucleus, which contains at least a vowel.
$Cd$ denotes the Coda, which mostly contains consonants.
$T$ denotes lexical tone, a feature existing in many languages to distinguish different words \cite{klein2001cross}\cite{lee2008role}\cite{anyanwu2008fundamentals}.
The syllabic structure above is shared across most languages in the world \cite{kessler1997syllable}\cite{wals-12}\cite{blevins2006syllable}.

However, how consonants ($\mathscr{C}$) and vowels ($\mathscr{V}$) constitute Onset, Nucleus, Coda differs across languages.
For instance, in English, an Onset can be a consonant cluster, such as ``sn'', while no consonant cluster can be the Onset of a syllable in Vietnamese or Cantonese \cite{Gimson1970}\cite{nguyen1990}\cite{slobin1986}\cite{mcleod2007}.

\subsubsection{Lexical Tones}
\label{subsubsec:background_tones}
In tonal languages, pitch is used to distinguish the meaning of words which are phonetically identical \cite{lee2008role}\cite{mojalefa2007}.
This distinctive pitch level or contour is referred to as lexical tones \cite{yip2002}.
For instance, there are 6 distinct lexical tones in Vietnamese \cite{hoang1965viet}
and 6 distinct lexical tones in Cantonese \cite{cheung2007}.
Around 70\% of languages are tonal \cite{yip2002}, concentrating in Africa, East and Southeast Asia \cite{wals13}.
Each lexical tone is commonly encoded in phonetic representation with a number.
For example, consider two different Vietnamese words: \verb|b_< O 3| (cow) and \verb|b_< O 6| (bug).
The two words are represented in phonetic units using X-SAMPA notation \cite{xsampa2001}, and have the same Onset (\verb|b_<|) and Nucleus (\verb|O|), but are distinguished by the two different lexical tones (tone 3 and tone 6).

\subsubsection{Transliteration}
\label{subsubsec:background_translit}
In transliteration, a word in a source language is converted to a target language while preserving the acoustic phonetic properties of the source language as much as possible. However, the syllabic structures of the transliterated output might differ from the syllabic structures of the original word \cite{knight1998}\cite{haugen1950analysis}\cite{frawley2003international}.

New phonetic units can be inserted into the transliteration output to imitate the pronunciation of the original word.
When performing transliteration on words from a non-tonal language (e.g. English) to tonal languages, lexical tones need to be assigned to each syllable of the transliterated output. So far, only limited preliminary work has explored lexical tones in transliteration \cite{hoang1965viet}\cite{kwong2009}.

Another example of insertion in transliteration is the addition of new phonemes to the output.
For example, converting a consonant cluster from English to many languages involves the insertion of an additional nucleus after the first consonant of a consonant cluster.
This phenomenon is defined as epenthesis with the inserted nucleus usually being a ``schwa'' and observed in languages such as Vietnamese \cite{hoang1965viet}, Japanese \cite{masuda2010processing}\cite{kashiwagi2008american}, Cantonese \cite{silverman1992}\cite{yip1993cantonese}, and many more \cite{rose2006}\cite{uffmann2007}\cite{yun2012}.

Furthermore, certain phonemes of the source word might be deleted in the target language due to phonological constraints.
For example, fricatives occurring at the syllable-final position of an English word tends to be omitted (deleted)
in their corresponding counter-part in Vietnamese \cite{hoang1965viet}
and Cantonese \cite{chan2000english}.

\subsection{Machine Transliteration}
\subsubsection{Input and Aim}
Suppose we are given a training dataset consisting of pairs of words from a source language (e.g. English) and their corresponding transliteration versions in a target language (e.g. Vietnamese).
Each word of the source language, which we name as source word in short, is represented in orthographic form as a sequence of letters $\mathbf{f} = [f_1, f_2, ..., f_m]$.
For each corresponding word in the target language, we can generate a sequence of phonemes $\mathbf{e} = [e_1, e_2, ..., e_n]$, which is also organized into syllables according to the phonological rules of the target language.

For example, given the English word \textit{Manchester}, its corresponding transliteration in Vietnamese is ``\foreignlanguage{vietnam}{man chét sờ tơ}'', written in Vietnamese text. Syllables of a Vietnamese word are separated by a whitespace. Vietnamese lexical tones are denoted by diacritics above the nuclei of the syllables, except tone 1 is not represented in text form by any diacritic mark.

\subsubsection{Approaches}

Symbolic systems for machine transliteration encapsulates expert-defined rules for mapping graphemes of the source word to the target language, as well as handling the mismatches between the pronunciation of the source and target languages (such as the epenthesis of nuclei and deletion of sound).
In \cite{wan1998automatic}, an English-Chinese name transliteration algorithm was devised using predefined rules.
An English word was first divided into syllables, each syllable is further divided into sub-syllabic units.
The sub-syllabic units were mapped to Pinyin characters. The syllables in Pinyin were then mapped to Chinese characters.
The syllabification rules were deterministic and the phonetic mapping used a lookup table.
In \cite{fujii2001japanese}, predefined ``phonetically similar'' English-Japanese (katakana) letters were used to derive English-Japanese symbol pronunciation similarity matrices.
To transliterate a new English word to Japanese, the decoder looked for the path that maximizes the total similarity across all letters.
An English-Punjabi transliteration system of name entities was devised in \cite{bhalla2013rule}.
Rules were defined to syllabify a source English word, with each syllable subsequently being mapped to a Punjabi syllable.

Standard statistical transliteration models are alignment-based. During training, the model learns (1) the distribution of how the source letters $\mathbf{f}$ are \textit{aligned} to the target phonemes $\mathbf{e}$ and (2) the distribution of how the aligned source letters $\mathbf{f}$ are mapped to the target phonemes $\mathbf{e}$. During decoding, the model produces the sequence of target phonemes with the highest likelihood.

The early machine transliteration system requires an additional step to convert the source letters to source phonemes, essentially convert the problem into a  phoneme-to-phoneme conversion \cite{virga2003}\cite{abduljaleel2003statistical}\cite{knight1998}\cite{stalls1998translating}.
In \cite{knight1998}, transliteration was posed as generative process and the English phonemes - Japanese phonemes alignment was estimated with expectation maximization \cite{dempster1977maximum}.
The same approach was applied to transliterate English word to Arabic text \cite{stalls1998translating}.
An adaptation was made by attaching  the position of the vowel in a word (initial, medial, final) to each English vowel.
The introduction of such contextual information helps the model learn the mapping of English letters to the less deterministically phonetic Arabic letters.
In \cite{virga2003}, English-Chinese transliteration was also performed using phonemic representation in an intermediate step.
Instead of aligning English phonemes to Chinese phonemes directly, the English phonemes were aligned to sub-syllabic units (initial-final) of the Chinese word.

The second group of standard alignment-based machine transliteration systems transform source graphemes to target phonemes without an intermediate source graphemes to source phonemes step.
The English-Arabic transliteration system in \cite{abduljaleel2003statistical} performed two runs of alignment.
In the first run, English letters were aligned to the Arabic characters and the most often aligned n-grams of English letters were extracted for the second run.
In the second run, the selected English n-grams were realigned to the Arabic characters.
In \cite{meng2001generating}, an English-Chinese transliteration comprised of two steps, the first step used expert rules to append syllable nuclei to the sequence of English phonemes and the second step modeled the probabilities of mapping between English and Chinese phonemes.

The third group of standard statistical transliteration models uses a combination of methods.
For example, the transliteration system for Arabic in \cite{al2002machine} combined both phonetic-based and letter-based transliteration.
The phonetic-based approach made use of the positional information of the vowel as in \cite{stalls1998translating}.
In \cite{oh2006ensemble}, English-to-Korean transliteration was performed by combining the output from an ensemble of grapheme-based transliteration model, phoneme-based transliteration model, and grapheme- and phoneme-based transliteration model.
The final output was produced by ranking with web data and relevance scores given by each transliteration model.

The joint source-channel model introduced in \cite{haizhou2004} (with a similar approach implemented in \cite{bisani2008}) approaches statistical transliteration differently.
Under the joint source-channel model, both the alignment and source-channel symbol mapping were handled intrinsically at the same time using intermediate tokens (grapheme-phoneme cosegments) comprising of both the source letters and the target phonemes.
The joint source-channel model has been shown to improve the transliteration performance in various tasks (e.g. \cite{cao2010}\cite{nguyen2016regulating}) and is used as the baseline for statistical transliteration in this work.

Recently, neural approaches have gained popularity in machine translation and other natural language processing tasks. End-to-end deep learning models are different from statistical transliteration models as they do not require explicit alignment between the source graphemes and the target graphemes.
However, among the few work \cite{rosca2016sequence, finch2016} to date that applied neural approaches to machine transliteration, none have shown that they outperform standard statistical approaches.

\subsubsection{Limitations of statistical approach}
\begin{figure}[!t]
\centering
\includegraphics[width=\linewidth]{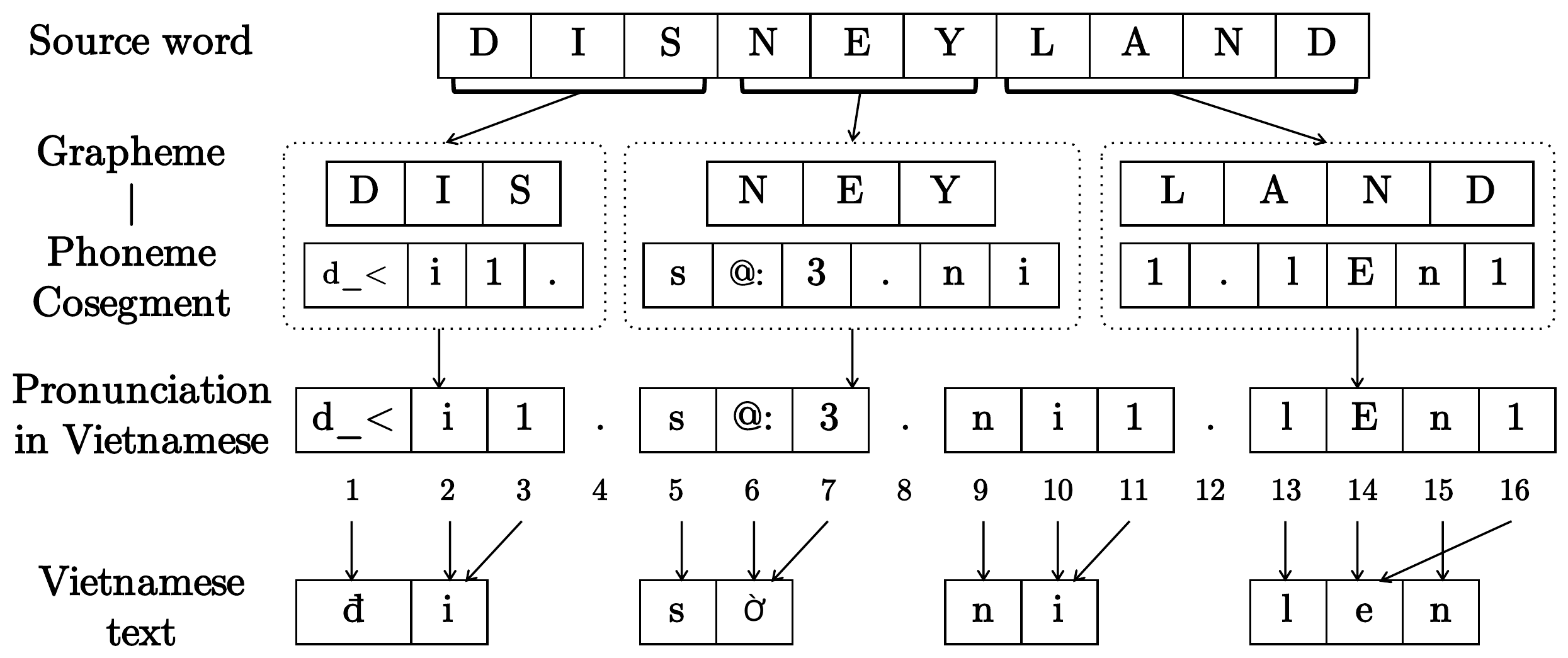}
\caption{Transliteration by the standard statistical transliteration model (no explicit phonological knowledge)}
\label{fig:joint_source_channel_output}
\end{figure}

Figure \ref{fig:joint_source_channel_output} illustrates how transliteration is performed via the alignment of grapheme-phoneme cosegments under the joint source-channel model.
\begin{figure}[t]
\centering
\includegraphics[width=\linewidth]{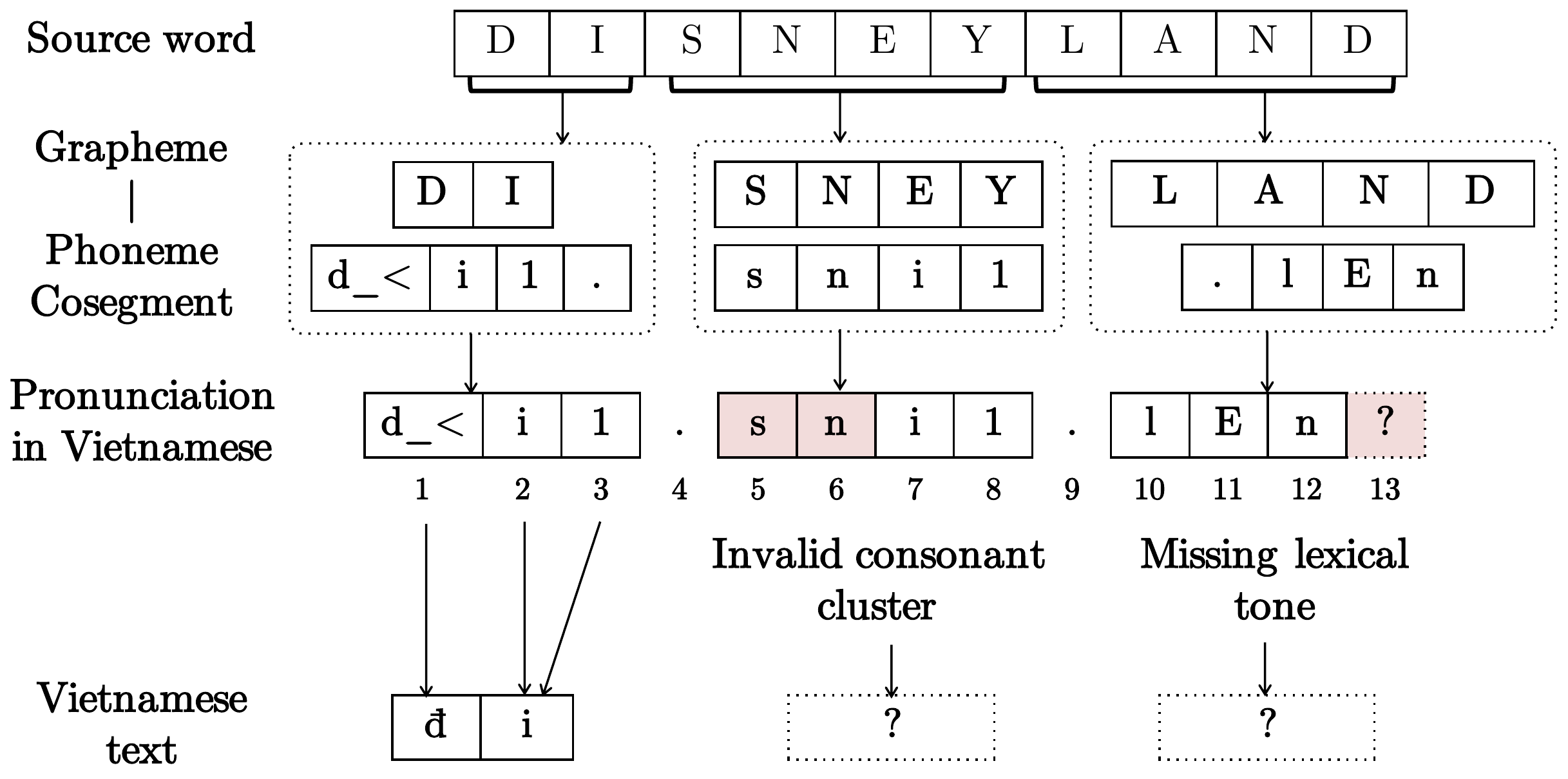}
\caption{Phonologically invalid transliteration output by the standard statistical transliteration model.}
\label{fig:joint_source_channel_invalid_output}
\end{figure}
In this example, the English word \textit{DISNEYLAND} is transliterated to Vietnamese.
The Vietnamese pronunciation is represented by X-SAMPA symbols.

Each dotted box represents a cosegment of source word's graphemes and target pronunciation's phonemes.
The arrows show the correspondence between the sequences of the English graphemes and sequences of the Vietnamese phonetic tokens via their cosegments.
The grapheme-phoneme cosegments in Figure \ref{fig:joint_source_channel_output} are the most likely sequence of cosegments given the input graphemes,
with the cosegments and their probability distribution learned from a training dataset.
The transliteration output is the sequence of phonetic tokens extracted from the most likely sequence of grapheme-phoneme cosegments.
The numbers of tokens $3$, $7$, $11$ and $16$ denote Vietnamese lexical tones.
The dot ``.'' of tokens $4$, $8$ and $12$ are delimiters between syllables.
The Vietnamese pronunciation can be mapped to Vietnamese text if the syllables conform to Vietnamese syllables' structure\footnote{The grapheme to phoneme mapping in Vietnamese is virtually one-to-one, and lexical tones are embedded in diacritics in vowels. To more explicitly explain the motivation and design philosophy of the transliteration framework, we choose to use the X-SAMPA phonetic symbols to represent the Vietnamese graphemes. The lexical tones are represented as diacritics that are above the graphemes, but in X-SAMPA format lexical tones are represented numerically, making it easier to explain the model.}.

In the standard statistical model for transliteration, phones, tones and syllabic delimiters of the target phonetic output are treated equally.
Since there is no phonological constraint placed on the organization of the components of the target pronunciation, transliteration is not guaranteed to produce outputs with structures considered valid by the phonological rules of the target language.
Figure \ref{fig:joint_source_channel_invalid_output} illustrates a scenario of which the standard transliteration model produces an output that is invalid according Vietnamese phonology:
\begin{itemize}
  \item In the second syllable, there is a consonant cluster \verb|sn| formed by phones at position $5$ and $6$. However, there is no consonant cluster in Vietnamese phonology\cite{nguyen1990}.
  \item In the third syllable, there is no lexical tone, while under Vietnamese phonology, each syllable is assigned with one tone\cite{hoang1965viet}.
\end{itemize}
Empirically, at least 21\% of transliteration outputs lacked lexical tones when we ran Sequitur \cite{g2p} (an implementation of the joint source-channel model) on 100 English - Vietnamese word-pronunciation pairs extracted from NIST OpenKWS13 corpus \cite{openkws2013}.

\section{Phonology-Augmented Statistical Model for Transliteration}
To overcome the limitations of the traditional statistical transliteration model, we attempt to augment statistical transliteration with phonological knowledge.
While many symbolic systems have been devised to capture the phonological rules involved in transliteration \cite{fujii2001japanese}\cite{ngo2014minimal}\cite{wan1998automatic}\cite{bhalla2013rule}, predefined rules are likely to make mistakes with words not observed in the training data.
Symbolic approaches are thus outperformed by statistical models in larger data sets \cite{ngo2014minimal}.
On the other hand, while statistical transliteration approaches like the joint source-channel model \cite{haizhou2004}\cite{bisani2008} can capture the phonological intricacies in converting a source word to a phonetically equivalent in another language, such approaches require a relatively large amount of training data to be effective \cite{ngo2014minimal}.

To overcome the limitations of the traditional statistical transliteration as well as the symbolic systems, we propose a phonology-augmented statistical model for transliteration by integrating phonological knowledge of the target language explicitly with a statistical model.
Phonological constraints have often been introduced to transliteration.
To adapt statistical machine translation tools into transliteration, many systems first converted the source written words into phonemes before performing alignment with the target phonemes \cite{virga2003}\cite{abduljaleel2003statistical}\cite{knight1998}\cite{stalls1998translating}.
The initial grapheme to phoneme conversion implicitly projected the source graphemes into phonetic units, which could be organized into sub-units of syllables \cite{virga2003}.
Some systems explicitly augmented the source word to improve the statistical model.
For example, in the English-to-Chinese transliteration system described in \cite{meng2001generating}, syllable nuclei were appended to the sequence of English phonemes to improve the accuracy of their conversion to Chinese.
In \cite{al2002machine}\cite{stalls1998translating}, attaching the position in a syllable (initial, medial, final) to each English vowel helped improve the mapping of English graphemes to Arabic graphemes.
In our proposed model, the positional information of a unit in a syllable (onset, nucleus, coda) are also used.
Unlike \cite{al2002machine}\cite{stalls1998translating}, such contextual information is used to inform the cross-language mapping of not just the nucleus, but also the onset and coda.
Imposing conditions on the structure of the output syllables helps to improve the phonological validity of the transliteration output, and yet remains generalizable across languages.
The same syllabic structure is shared across most languages \cite{wals-12}\cite{blevins2006syllable} and thus, utilizing this universal phonological property keeps the proposed framework relatively language-independent.
On the other hand, when the transliteration output conforms to valid syllabic structures, the search space for the output is also bounded, which might make finding the correct output easier.
For example, the consonant ``l'' in English can be mapped to different Vietnamese phonemes.
However, conditioned on the corresponding sub-syllabic unit that ``l'' is mapped to in the output syllable, the number of possibilities would be smaller:
``l'' is more likely to correspond to an \verb\/n\/ phoneme in Vietnamese if it is at the Coda position, and more likely to correspond to an \verb\/l\/ phoneme if it is at the Onset position.

Furthermore, transliteration is  performed directly from the source graphemes to the target phonemes.
By avoiding using the intermediate source phonemes, the proposed model does not assume the specific language of the source word.

\begin{figure}[t]
  \centering
    \centerline{\includegraphics[width=\linewidth]{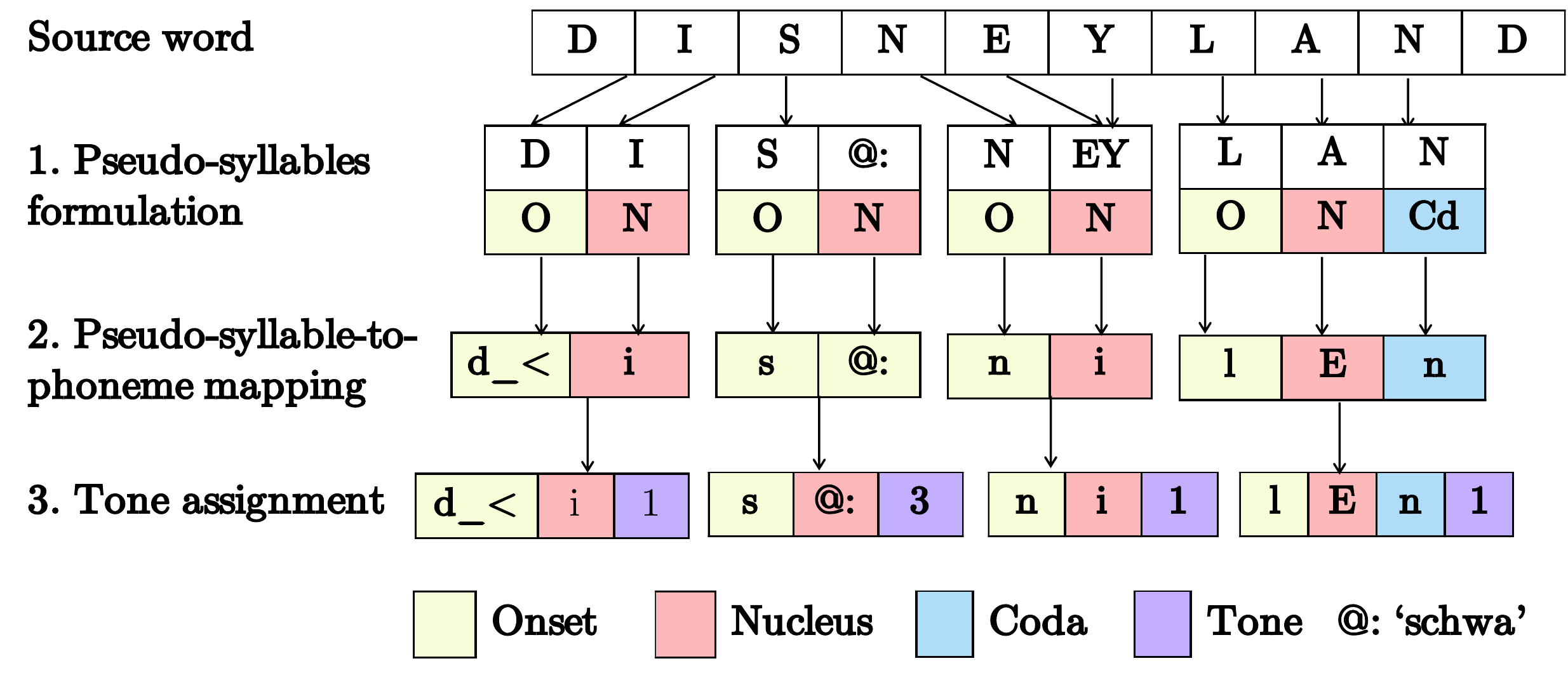}}
\caption{Phonology-augmented statistical model for transliteration}
\label{fig:proposed_translit}
\end{figure}

The proposed model performs translieration in three steps, as summarized in Figure \ref{fig:proposed_translit}.

\begin{enumerate}
  \item Pseudo-syllable formulation: graphemes of the source word are organized into pseudo-syllables.
  Source graphemes are assigned explicitly to the sub-syllabic units of each pseudo-syllable such that the units form valid syllabic structures defined by the target language's phonology.
  \item Pseudo-syllable-to-phoneme mapping: a language model is used to map the graphemes of each pseudo-syllable, given their assigned unit in a syllable, to the most likely phonemes.
  \item Tone assignment: one tone is assigned to each syllable, based on the target language's phonemes in each syllable.
\end{enumerate}

\subsection{Pseudo-Syllable Formulation}
\label{subsec:pseudo-syl_formulation}
\subsubsection{Scheme}
Pseudo-syllable is a representation of how segments of a foreign word are arranged according to the syllable structure specified by the target language's phonology.
The concept is inspired by how native speakers process a foreign loanword by imposing native phonological constraints on the foreign word's form \cite{knight1998}\cite{silverman1992}.
A pseudo-syllable's structure is defined as: $s_{i}=\{[{s}_{i}^O], {s}_{i}^N, [{s}_{i}^{Cd}]\}$, where $s_{i}^{O}$, $s_{i}^{N}$, $s_{i}^{Cd}$ are the Onset, Nucleus and Coda of the $i$-th pseudo-syllable respectively.
Onset, Nucleus and Coda are sub-syllabic units constituting a syllable.\\

\noindent\textit{i. Graphemic labels assignment}

To form pseudo-syllables from a source word, a segmentation function groups the source word's graphemes into sub-syllabic units.
Pseudo-syllables are formulated by grouping graphemes of the source word according to the labels that they are assigned to.
The labels suggest which sub-syllabic unit the grapheme would be assigned to, and the grapheme's relation with neighboring graphemes.
For illustration, we will use five labels \verb|O, N, Cd, ON, X| in our discussion:
\begin{itemize}
  \item \verb|O|: \textbf{O}nset
  \item \verb|N|: \textbf{N}ucleus
  \item \verb|Cd|: \textbf{C}o\textbf{d}a
  \item \verb|ON|: \textbf{O}nset, with a special token representing a new \textbf{N}ucleus inserted to the pseudo-syllable containing this Onset.
  This label addresses the insertion phenomenon explained in Section \ref{subsubsec:background_translit}.
  \item \verb|X|: e\textbf{X}cluded. Graphemes labelled by \verb|X| are not assigned to any pseudo-syllable.
  This label addresses the deletion phenomenon explained in Section \ref{subsubsec:background_translit}.
\end{itemize}
The sequence of labels assigned to all graphemes of a source word is the output of the pseudo-syllabic formulation step.\\
\begin{figure}[t]
  \centering
    \centerline{\includegraphics[width=\linewidth]{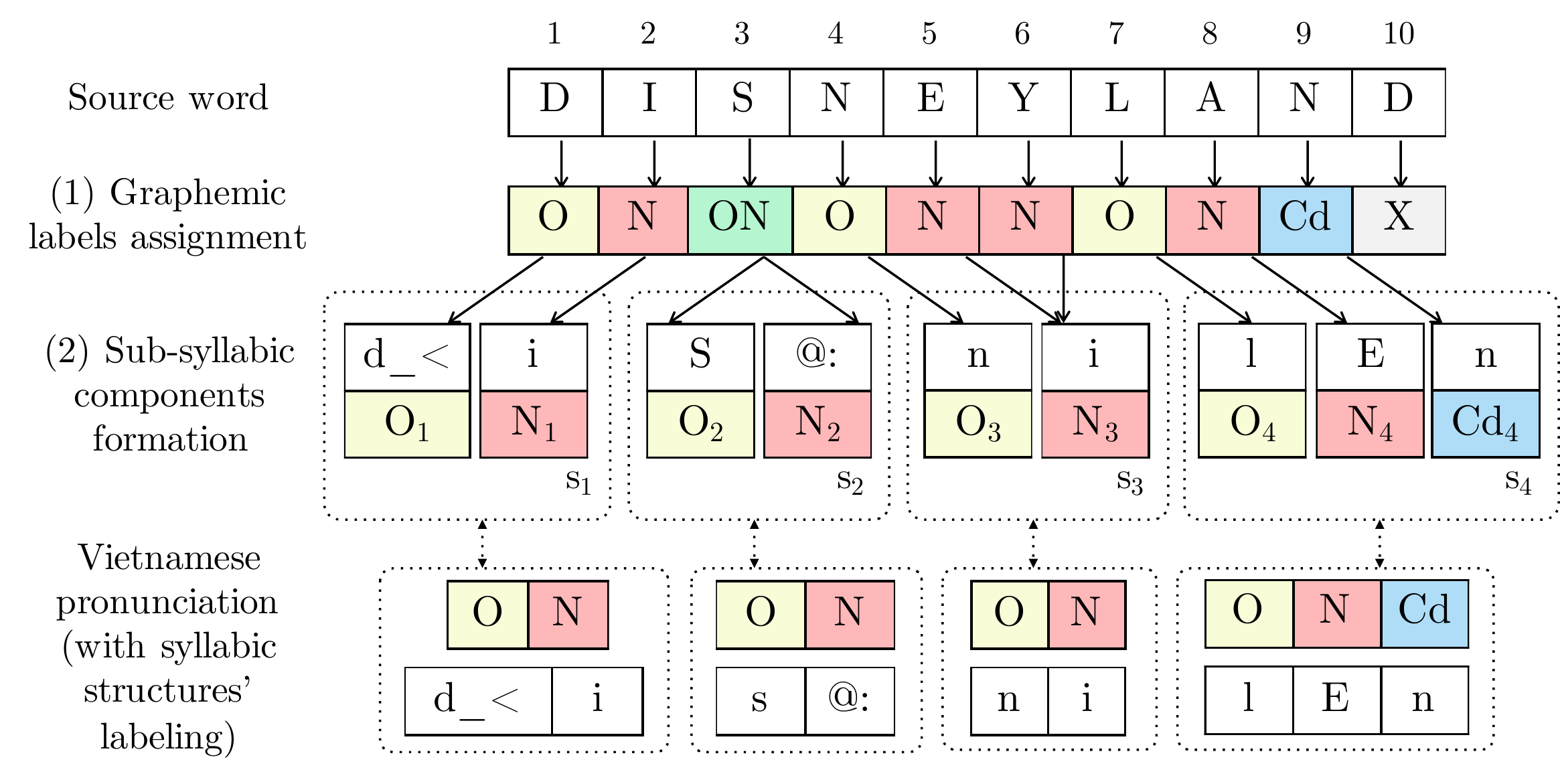}}
\caption{Pseudo-syllable formulation}
\label{fig:pseudo_syllabic_labels_demo}
\end{figure}

\noindent\textit{ii. Sub-syllabic units formation}
\subsubsection{Ground-truth}
\label{subsubsec:pseudo-syl_groundtruth}

To train a model for pseudo-syllabic formulation, a ground-truth of graphemic labels need to be determined from the original training pairs of original source words and target pronunciation.
For a given training pair, a search is performed among all possible combinations of graphemic labels in order to find the candidate combination that produces a sequence of pseudo-syllables of the same structures as the target pronunciation's.
To reduce the computational complexity of the search, any partial sequence of graphemic labels that produces invalid pseudo-syllables (for example, pseudo-syllables that start with a coda) would be rejected without iterating through the remaining graphemes.
Step 1 of Figure \ref{fig:pseudo_syllabic_labels_demo} shows an example of a combination of graphemic labels.
Given the sequence of graphemic labels in the figure, the corresponding pseudo-syllables are formed as follows:
\begin{itemize}
  \item Labels at position 1, 4 and 7 are \verb|O|. Therefore, the corresponding graphemes `D', `N' and `L' are assigned to the Onsets of pseudo-syllables $s_{1}$, $s_{3}$, and $s_4$ respectively.
  \item Labels at position 2, 5, 6 and 8 are \verb|N|. Therefore, the corresponding graphemes are assigned to the Nuclei of the pseudo-syllables.
   Note that since the graphemes `E' and `Y' at position 5 and 6 are adjacent and both given label \verb|N|, they are joined to form the Nucleus of pseudo-syllable $s_{3}$.
  \item Label at position 3 is \verb|ON|. Therefore, the corresponding grapheme `S' is assigned to the Onset of pseudo-syllable $s_{1}$.
    A special token ($@:$) is also inserted to the Nucleus position of pseudo-syllables $s_{1}$, representing an epenthesized nucleus. 
  \item Label at position 10 is \verb|X|. Therefore, the corresponding grapheme `D` is ignored.
\end{itemize}
The resulting pseudo-syllables of the word \textit{DISNEYLAND} are shown in Figure \ref{fig:pseudo_syllabic_labels_demo} as dotted boxes, with:
\begin{itemize}
  \item $s_{1}$=\{D, I\}, where $s^{O}_{1}$=\{D\}, $s^{N}_{1}$=\{I\}
  \item $s_{2}$=\{S, @:\}, where $s^{O}_{2}$=\{S\}, $s^{N}_{2}$=\{@:\}
  \item $s_{3}$=\{N, EY\} where $s^{O}_{3}$=\{N\}, $s^{N}_{3}$=\{EY\}
  \item $s_{4}$=\{L, A, N\} with $s^{O}_{4}$=\{L\}, $s^{N}_{4}$=\{A\}, and $s^{Cd}_{4}$=\{N\}
\end{itemize}
The graphemic labels in Figure \ref{fig:pseudo_syllabic_labels_demo} shows a valid pseudo-syllable formulation since the pseudo-syllables have the same syllabic structures \verb|O N . O N . O N .O N Cd| as the syllables of the target pronunciation.\\

\subsubsection{Training}
\label{subsubsec:pseudo-syl_training}

Given a training example consisting of a pair of foreign word $\mathbf{f}=[f_{1}, f_{2},..., {f}_{M}]$ and its corresponding phonetic output $\mathbf{e}=[e_{1}, e_{2},..., {e}_{N}]$, the corresponding training example for graphemic labels can be deduced as follows:

Let $\mathbf{l} = [l_{1}, l_{2}, ..., l_{M}]$ be the sequence of graphemic labels assigned to the graphemes sequence $f$. For $1\leq m \leq M$, $l_{m} \in \mathcal{L}$ where $\mathcal{L}$ is the set of all possible graphemic labels, $\mathcal{L} = \{\verb|O, N, Cd, ON, X|\}$.

Let $\Gamma(\mathbf{f}, \mathbf{l})$ be the function to form sub-syllabic units from the sequence of graphemes $\mathbf{f}$ and sequence of graphemic labels $\mathbf{l}$ as described in Section \ref{subsec:pseudo-syl_formulation}-ii.
Function $\Gamma$ is defined as:
\begin{align}
  \Gamma(\mathbf{f}, \mathbf{l})
  &=\mathbf{s}=[s_{1}, s_{2}, ..., s_{K}] \\
  &=[(s_{1}^{O}, s_{1}^{N}, s_{1}^{Cd}), ..., (s_{K}^{O}, s_{K}^{N}, s_{K}^{Cd})]
\label{eq:syl_T_mle}
\end{align}
\noindent where $\mathbf{s}=[s_{1}, s_{2}, ..., s_{K}]$ is a sequence of pseudo-syllables, with $(s_{k}^{O}, s_{k}^{N}, s_{k}^{Cd})$ being the groups of graphemes assigned to the Onset, Nucleus and Coda unit of the $k$-th pseudo-syllable respectively.

A sequence of graphemic labels is found for $\mathbf{f}$ and $\mathbf{e}$ if the syllabic structures $\mathbf{r}_s$ of the pseudo-syllables $\mathbf{s}$ match the syllabic structures $\mathbf{r}_e$ of the target pronunciation $\mathbf{e}$.
For example, in Figure \ref{fig:pseudo_syllabic_labels_demo}, the syllabic structures $\mathbf{r}_s$ and $\mathbf{r}_e$ are both [$O, N$], [$O, N$], [$O, N$], [$O, N, Cd$].
The syllabic structures $\mathbf{r}_e$ of the target pronunciation $\mathbf{e}$ can be trivially determined using a pronunciation-subsyllabic unit dictionary of the target language.
For example, in the case of target transliteration output for Vietnamese, the language specific document of the OpenKWS13 is used as the pronunciation-subsyllabic unit dictionary \cite{openkws2013}.
The syllabic structures $\mathbf{r}_s$ can be directly determined from the pseudo-syllables $\mathbf{s}$.\\

\subsubsection{Decoding}
\label{subsubsec:pseudo-syl_decoding}

We want to find the most likely graphemic roles $\mathbf{l}$ for all letters of the new example $\mathbf{f}$:
\vspace{-0.2cm}
\begin{eqnarray}
  \mathbf{l}^{*}
  &=\operatorname*{arg\,max}_{\mathbf{l}}p\left(\mathbf{l}\mid\mathbf{f}, \mathcal{D}\right)
\label{eq:graphemic_labels_mle}
\end{eqnarray}

\noindent where $\mathcal{D}$ is the training data of graphemic labels produced in the previous section.

We model $\mathbf{l}$ as a Markov chain of n-gram, with n being an even number:
\vspace{-0.2cm}
\begin{eqnarray}
  \mathbf{l}^{*}
  &=\operatorname*{arg\,max}_{\mathbf{l}}\prod_{i=1}^{M}p\left({l}_{i}\mid\left({f}_{i-n/2}, ..., {f}_{i+n/2}\right)\right)
\label{eq:graphemic_labels_ngram}
\end{eqnarray}

\subsubsection{Smoothing}
\label{subsubsec:pseudo-syl_smoothing}

To deal with data sparsity, ${l}_{i}$ in Eq. (\ref{eq:graphemic_labels_ngram}) is estimated from a weighted score of probability of smoothed n-grams:
\begin{eqnarray}
  q\left({l}_{i}\right)
  &=\sum_{k=0}^{n/2}\sum_{t=0}^{n/2}\omega_{k, t}p\left({l}_{i}\mid\left({f}_{i-n/2}',...,{f}_{i+n/2}'\right)\right)
\label{eq:syl_T_mle}
\end{eqnarray}
\noindent where $(f_{i-n/2}', ..., f_{i+n/2}')$ is a smoothed n-gram such that given $0\leq{k, t}\leq{n/2}$ ($n$ is an even number):

\vspace{0.3cm}
$f_{j}'=\begin{cases}
f_{j} \mbox{~~if $i-n/2+t\leq{j}\leq{i+n/2-k}$} \\
\text{otherwise} \begin{cases} \mathscr{C} &\mbox{if $f_{j}$ is a consonant}  \\
                  \mathscr{V} &\mbox{if $f_{j}$ is a vowel}      \\
                  $\_$ &\mbox{if $f_{j}$ is $\mathscr{C}$ or $\mathscr{V}$}      \\
                 \end{cases}
\end{cases}$
\vspace{0.3cm}

\noindent where the empty token $\_$ represents the back-off case of which the smoothed n-gram is the shortened version of the original n-gram. For example, given the tri-gram of letters (B, E, S), its smoothed n-grams are \{(B, E, S), ($\mathscr{C}$, E, S), (B, E, $\mathscr{C}$), ($\mathscr{C}$, E, $\mathscr{C}$), ..., ($\mathscr{V}$)\}. $\omega_{k, t}$ is the weight corresponding to the smoothed n-gram, with $\omega_{k, t}>\omega_{k', t'}$ for $k+t<k'+t'$; for example: $\omega_{1,0}$ (corresponding to ($\mathscr{C}, \text{E}, \text{S}$)) $> \omega_{1,1}$ (corresponding to ($\mathscr{C}, \text{E}, \mathscr{C}$)).\\

\subsection{Pseudosyllable-to-Phoneme Mapping}
\label{subsec:g2p_mapping}
Transliteration takes into account (1) graphemes of the source word, (2) how the word is pronounced in the source language, and (3) phonological rules of the target language \cite{hoang1965viet}\cite{silverman1992}.
A formal equation of such relation can be given by:
\vspace{-0.2cm}
\begin{eqnarray}
  \mathbf{e}^{*}
  =\operatorname*{arg\,max}_{\mathbf{e}}p\left(\mathbf{e}\mid\mathbf{f},\mathbf{v},\mathcal{L}_{e}\right)
  \label{eq:phone_mle}
\end{eqnarray}
\vspace{-0.4cm}

\noindent where $\mathbf{e}$ is the phonemes of the target language's pronunciation, $\mathbf{f}$ and $\mathbf{v}$ are the graphemes and phonemes of the original word from the source language, and $\mathcal{L}_{e}$ is the phonological rules of the target language.

In the proposed model, the relationship in Eq. (\ref{eq:phone_mle}) is simplified to:
\vspace{-0.2cm}
\begin{eqnarray}
  \mathbf{e}^{*} &= \prod_{k=1}^{K}\prod_{u\in \mathcal{U}}\operatorname*{arg\,max}_{e_{k}^{u}}p\left(e_{k}^{u}\mid~s_{k}^{u}, v_{k}^{u}\right)
  \label{eq:syl_unit_mle}
\end{eqnarray}
\vspace{-0.4cm}

\noindent with $e_{k}^{u}$, $s_{k}^{u}$, $v_{k}^{u}$ being the target language's phoneme,
source language's graphemes and source language's phoneme of the sub-syllabic unit $u\in\mathcal{U}=\{O, N, Cd\}$,
of the $k$-th syllable in the target pronunciation or the $k$-th pseudo-syllable of the source word.

The source phonemes are produced with CMU text-to-phoneme tool for American English \cite{cmu_t2p}.
Note that we do not assume that the source words are in English, the source phonemes are only used as back-off to complement the source graphemes in performing the pseudo-syllables to phonemes mapping.

\begin{figure}[t]
  \centering
    \centerline{\includegraphics[width=\linewidth]{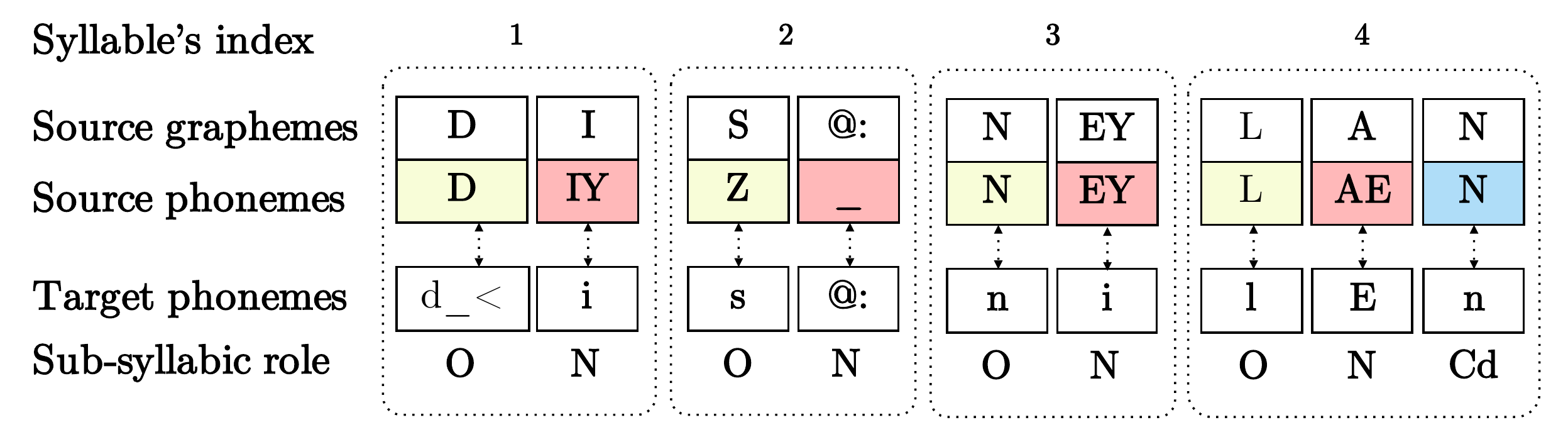}}
\caption{Pseudo-syllable-to-Phoneme mapping with phonological constraint}
\label{fig:grapheme_to_phoneme}
\end{figure}

The mapping from pseudo-syllables to phonetic tokens in the proposed model is illustrated in Figure \ref{fig:grapheme_to_phoneme}.
As seen from Figure \ref{fig:grapheme_to_phoneme}, pseudo-syllables formulated in Section \ref{subsec:pseudo-syl_formulation} provide a streamlined model of the phonological constraints $\mathcal{L}_{e}$ by modeling 
(1) valid syllabic structures in the target pronunciation, 
and (2) valid phonemes for each sub-syllable unit $r$.
The distribution $p\left(\mathrm{e}_{k}^{u}\mid\mathrm{s}_{k}^{u},\mathrm{v}_{k}^{u}\right)$ of Eq. \ref{eq:syl_unit_mle} can be learned from the training data prepared in Section \ref{subsec:pseudo-syl_formulation}.

\subsection{Lexical Tone Assignment}
\label{subsubsec:tone_assignment}
In most transliteration models, lexical tones are treated the same as other phonetic tokens of the transliterated output.
Lexical tone assignment usually depends on large amounts of training data to be correct \cite{virga2003}\cite{haizhou2004}.
Previous transliteration studies using tone assignment to complement output of statistical models have shown improvement in transliteration performance \cite{kwong2009}\cite{song2010}.

Phonology studies show that in many tonal languages, the assignment of a lexical tone to a syllable is influenced by the syllable's phonemes \cite{hoang1965viet}\cite{setter2010}, and the lexical tones of adjacent syllables (tone sandhi) \cite{lee1997}\cite{takenobu2008}.
Similarly, we used tonal and phonetic context to model tone assignment to a syllable.

Let $\mathbf{t}$ be the lexical tones assigned to all syllables of a transliteration output, with ${t}_{k}$ being the lexical tone assigned to the $k$-th syllable of the output.
\vspace{-0.1cm}
\begin{eqnarray}
  \mathbf{t} = \operatorname*{arg\,max}_{\mathbf{t}}\prod_{k}p\left(t_{k}\mid{t}_{k-1},(\mathrm{e}_{k}^{O}, \mathrm{e}_{k}^{N}, \mathrm{e}_{k}^{Cd}),t_{k+1}\right)
  \label{eq:sim_tones_mle}
\end{eqnarray}

\noindent where $(\mathrm{e}_{k}^{O}, \mathrm{e}_{k}^{N}, \mathrm{e}_{k}^{Cd})$ are phonemes of the Onset, Nucleus, and Coda sub-syllabic unit in the $k$-th syllable of the output.

\section{Experiments}
\label{sec:exp}
\subsection{Transliteration for different language pairs}
\subsubsection{Experimental set-up}
We performed transliteration on corpora of two language pairs: English-to-Vietnamese and English-to-Cantonese\footnote{It might be more linguistically precise to say Western-to-Vietnamese and Western-to-Cantonese, given some of the English words are not of Anglophone origin.}.

  a) English-to-Vietnamese Corpus from NIST OpenKWS13 Evaluation:
The Vietnamese corpus is released from the IARPA Babel program \cite{openkws2013} for the NIST OpenKWS13 Evaluation (denoted as NIST OpenKWS13 dataset).
From the scripted telephone speech of the OpenKWS13 lexicon, foreign words in the developmental data (\textit{evalpart1}, which was publicly released)  were designated as the test set (140 words) and shared among all experiments.
Training-development data sub-corpora of size 100, 200, 300, 400, 500, 587 were also extracted.
Each sub-corpus is split by the proportion of 75\% to 25\% into training and development sets.
Each sub-corpus is repartitioned 4 times to create 4 different, non-overlapping training-development data sets.
Each phonetic output of the foreign words is a sequence of Vietnamese phonemes in X-SAMPA \cite{xsampa2001} and one lexical tone (represented by a number from 1 to 6), separated by syllabic delimiters.
The setup is described in details in \cite{ngo2014minimal}.

  b) English-to-Cantonese Corpus from Hong Kong Polytechnic University:
This corpus comprises of 473 pairs of foreign words and their corresponding Cantonese pronunciations, extracted from a database of English loanwords in Hong Kong Cantonese, developed at the Hong Kong Polytechnic University \cite{wong2009integration} (denoted as Hong Kong Polytechnic Cantonese dataset).

Each pronunciation is a sequence of syllables of Cantonese in Jyutping \cite{jyutping} and one lexical tone (represented by a number from 1 to 6), separated by syllabic delimiters\footnote{For Cantonese, we chose to use Jyuping (phoneme-baed graphemes) instead of logographic graphemes due to the following reasons: (1) Jyuping explicitly represents lexical tones, while logographic graphemes usually do not embed lexical tone information; (2)  Logographic graphemes lack a standard system in Cantonese, especially for words specific to Cantonese and have no equivalent counterpart in Mandarin. We have a separate piece of work on pronunciation modeling for Han logographic languages under review.}.
Sub-corpora of size 100, 200, 300, 400 and 473 were randomly sampled for the experiments.
Each sub-corpus of the Cantonese data set was split randomly into three sets, 60\% for training the models, 20\% for development, and 20\% testing.
Each sub-corpus is repartitioned 5 times to create 5 different, non-overlapping test sets (cross-validation).

\subsubsection{Implementation Details}
In each experiment, transliteration was performed using three different approaches:

  a) Standard statistical approach (no phonological constraint): joint source-channel model implemented with Sequitur \cite{g2p}.

  b) Symbolic approach: two symbolic systems for Vietnamese and Cantonese were implemented for the respective transliteration experiments of the two languages.
The symbolic transliteration system used in these experiments was first proposed for Vietnamese language in \cite{ngo2014minimal} and further extended to Cantonese (Appendix \ref{app:rule_based}).
  The symbolic systems was optimized to the best of our capabilities to minimize the string error rate of the outputs.
  The error rates by the symbolic systems served as reference to compare the performance by the statistical models.

  c) Proposed phonology-augmented statistical model for transliteration.

\subsubsection{Evaluation Metrics}
The error rates computed in our experiments were evaluated using SCLITE \cite{sclite}:

  a) Token error rate (TER): Tokens include both phonemes and syllable delimiters.

  b) String error rate (SER): Any error within a string results in string error.
  
\subsection{Results}
  In all experiments on Vietnamese and Cantonese corpora, the phonology-augmented statistical framework improved upon the transliteration performance of the standard joint-source channel model under limited-resource scenarios.
  As training data sizes increased, the transliteration performance by statistical models improved and caught up with the symbolic system's performance.
  Transliteration error rates by the symbolic system for Vietnamese in Figures \ref{fig:kws_ter} - \ref{fig:kws_ser} (and also in Figure \ref{fig:kws_syler} - \ref{fig:kws_tone}) are flat across all sizes of the training data because the same test set was used.
  On the other hand, transliteration error rates by the symbolic system for Cantonese in Figures \ref{fig:cantonese_ter} - \ref{fig:cantonese_ser} (and also in Figure \ref{fig:hkpoly_syler} - \ref{fig:hkpoly_tone}) vary across different corpus sizes because some words that overlap across different folds of the cross-validation contributed multiple times to the error rates, even though the symbolic system would produce the same outputs for these input words.
  \subsubsection{English-to-Vietnamese}
  \label{subsubsec:vie_exp}
\begin{figure}[t]
  \centering
  \begin{subfigure}[b]{0.3\textwidth}
    \centering
    \caption{Token error rate}
    \vspace{-0.2cm}
    \centerline{\includegraphics[height=5cm]{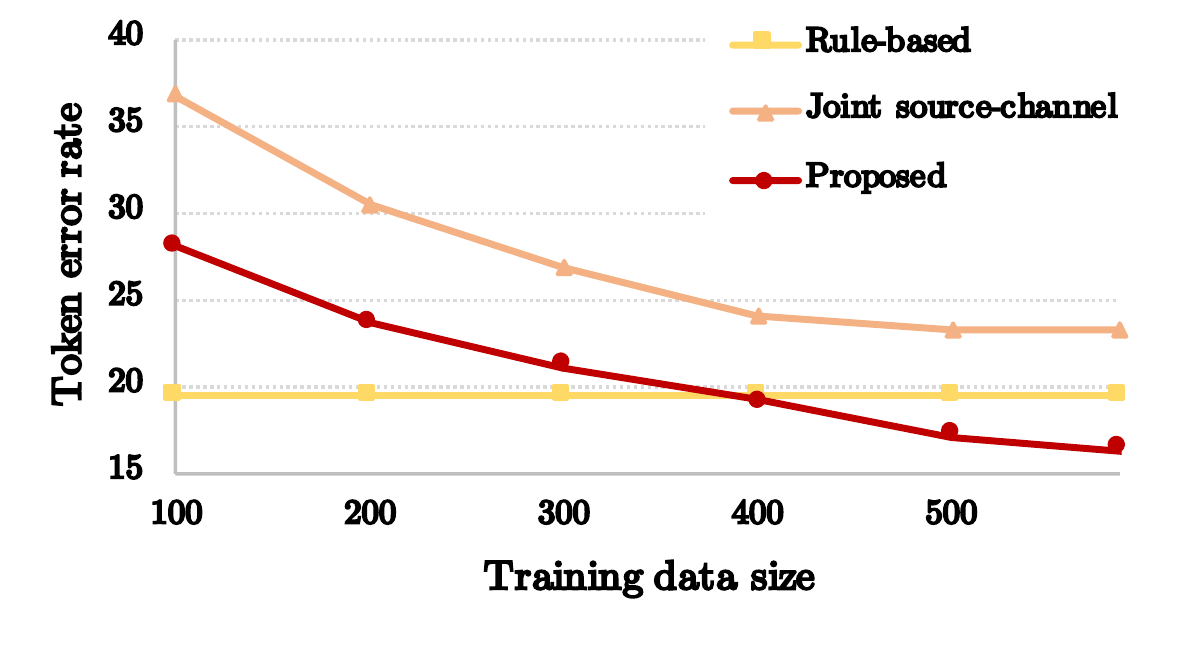}}
    \vspace{-0.2cm}
    \label{fig:kws_ter}
  \end{subfigure}
  \begin{subfigure}[t]{0.3\textwidth}
    \centering
    \caption{String error rate}
    \vspace{-0.2cm}
    \centerline{\includegraphics[height=5cm]{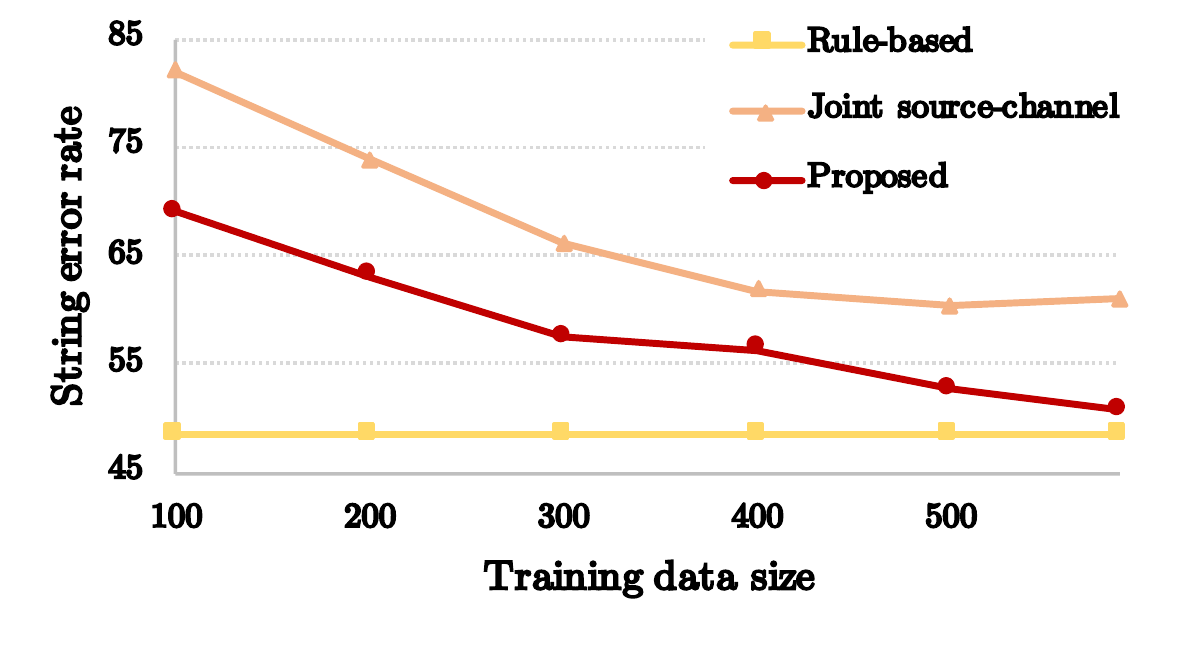}}
    \label{fig:kws_ser}
  \end{subfigure}
  \caption{Performance of different transliteration models as a function of corpus size (Vietnamese dataset - NIST OpenKWS13 dataset).}    
  \label{fig:vie_performance}
\end{figure}

   From Figures \ref{fig:kws_ter} - \ref{fig:kws_ser}, we see that the proposed model consistently outperforms the statistical baseline across all the training set sizes for English-to-Vietnamese.

   The proposed model outperforms the baselines at the smallest training data size (100 word pairs): it improves the joint source-channel model by 23.71\% relative in TER and by 15.85\% relative in SER.
   The proposed model outperforms the baselines at the largest training data size (587 word pairs):
     it improves the joint source-channel model by 29.74\% relative in TER
     and by 16.89\% relative in SER.
    The proposed model also improves the symbolic system by 15.98\% relative in TER.

  \subsubsection{English-to-Cantonese}
  \label{subsubsec:cantonese_exp}
\begin{figure}[th]
  \centering
  \begin{subfigure}[b]{0.3\textwidth}
    \centering
    \caption{Token error rate}
    \centerline{\includegraphics[height=5cm]{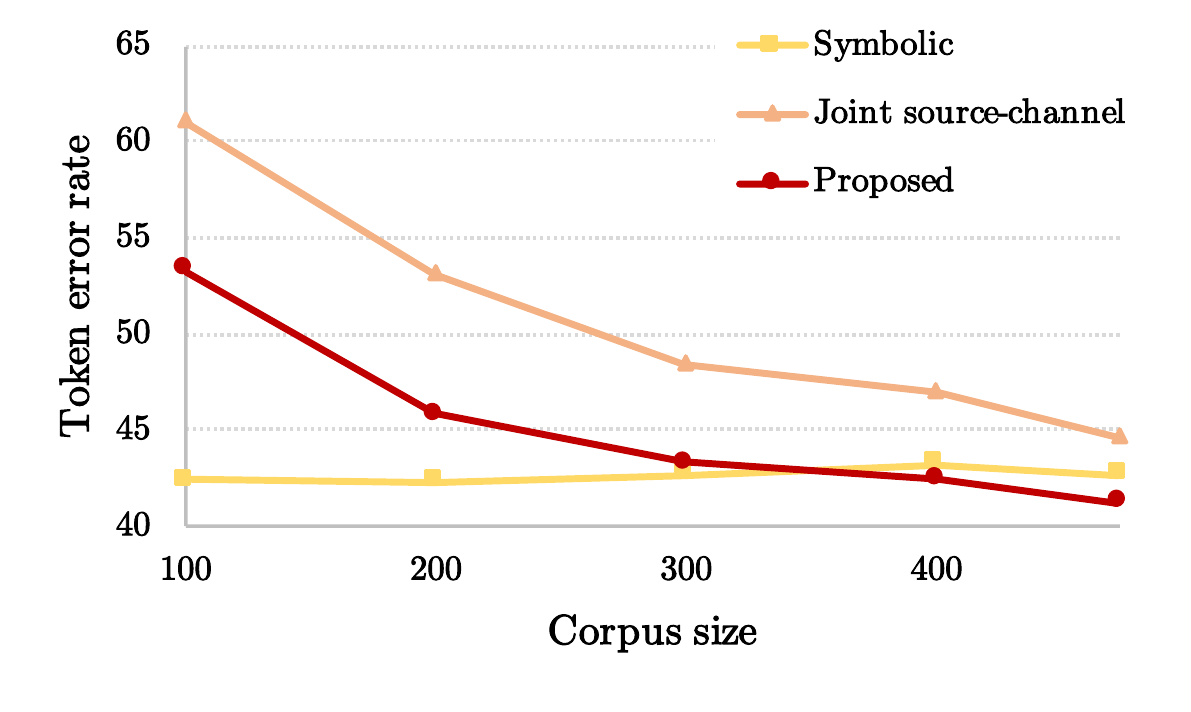}}
    \label{fig:cantonese_ter}
    \vspace{-0.2cm}
  \end{subfigure}
  \begin{subfigure}{0.3\textwidth}
    \centering
    \caption{String error rate}
    \vspace{-0.2cm}
    \centerline{\includegraphics[height=5cm]{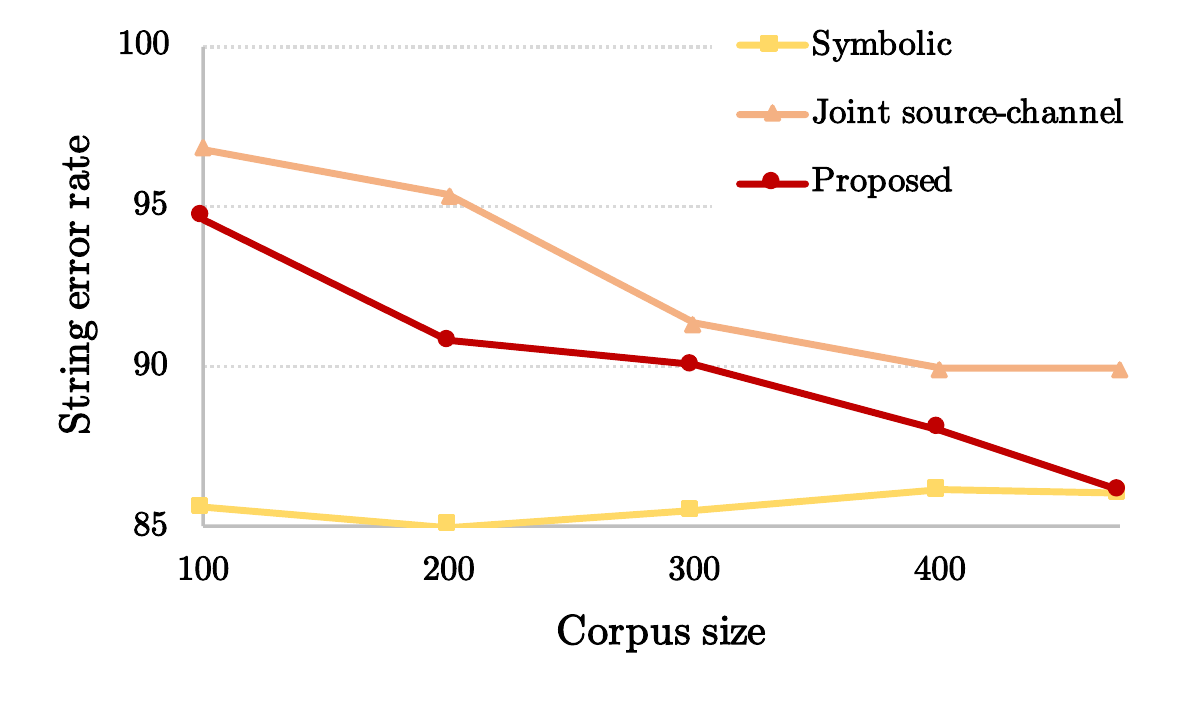}}
    \label{fig:cantonese_ser}
  \end{subfigure}
  \caption{Performance of different transliteration models as a function of corpus size (Hong Kong Polytechnic Cantonese dataset).}    
  \label{fig:cantonese_performance}
\end{figure}
From Figures \ref{fig:cantonese_ter} - \ref{fig:cantonese_ser}, we observe that the proposed model outperforms the statistical baseline across all the training set sizes for English-to-Cantonese.

The proposed model outperforms the baselines at the smallest corpus size (100 word pairs):
  it improves the joint source-channel model by 12.77\% relative in TER
  and by 2.27\% relative in SER.
  
The proposed system outperforms the statistical baseline at the largest corpus size (473 word pairs)
  by 7.83\% relative in TER and by 4.23\% relative in SER.
The proposed model also improves the the symbolic system by 3.51\% relative in TER.

\begin{figure}[ht]
  \centering
    \begin{subfigure}[b]{0.3\textwidth}
    \centering
    \caption{Syllable error rate}
    \vspace{-0.2cm}
    \centerline{\includegraphics[height=4.2cm]{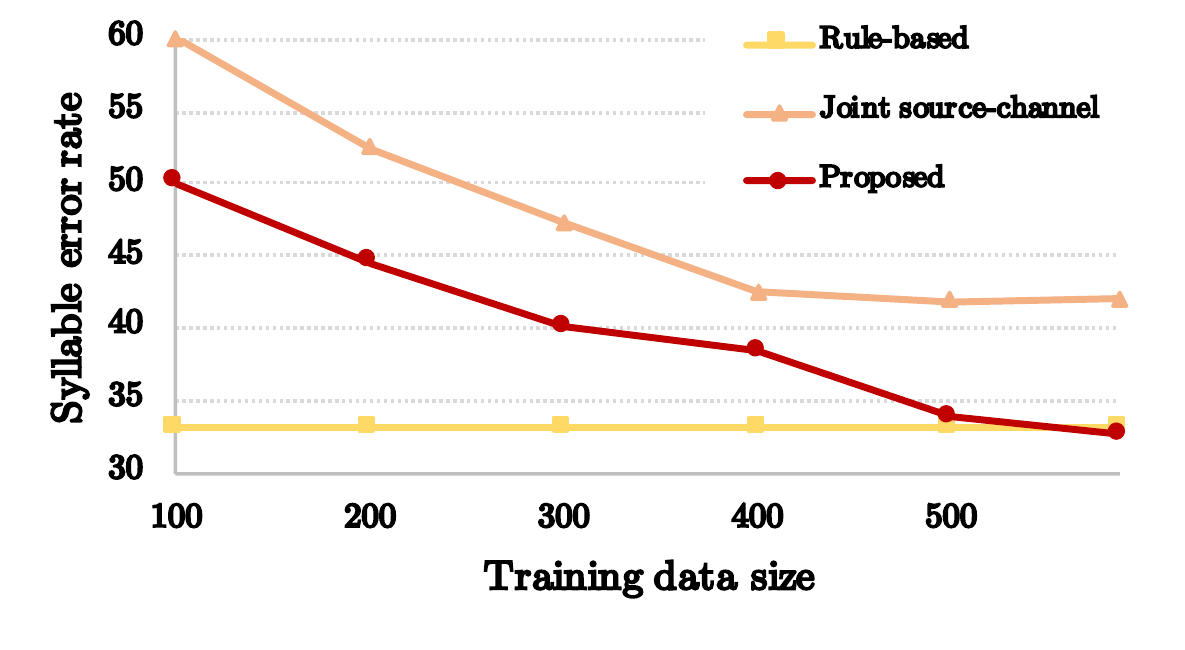}}
    \label{fig:kws_syler}
  \end{subfigure}
  \begin{subfigure}[b]{0.3\textwidth}
    \centering
    \caption{Onset error rate}
    \vspace{-0.2cm}
    \centerline{\includegraphics[height=4.2cm]{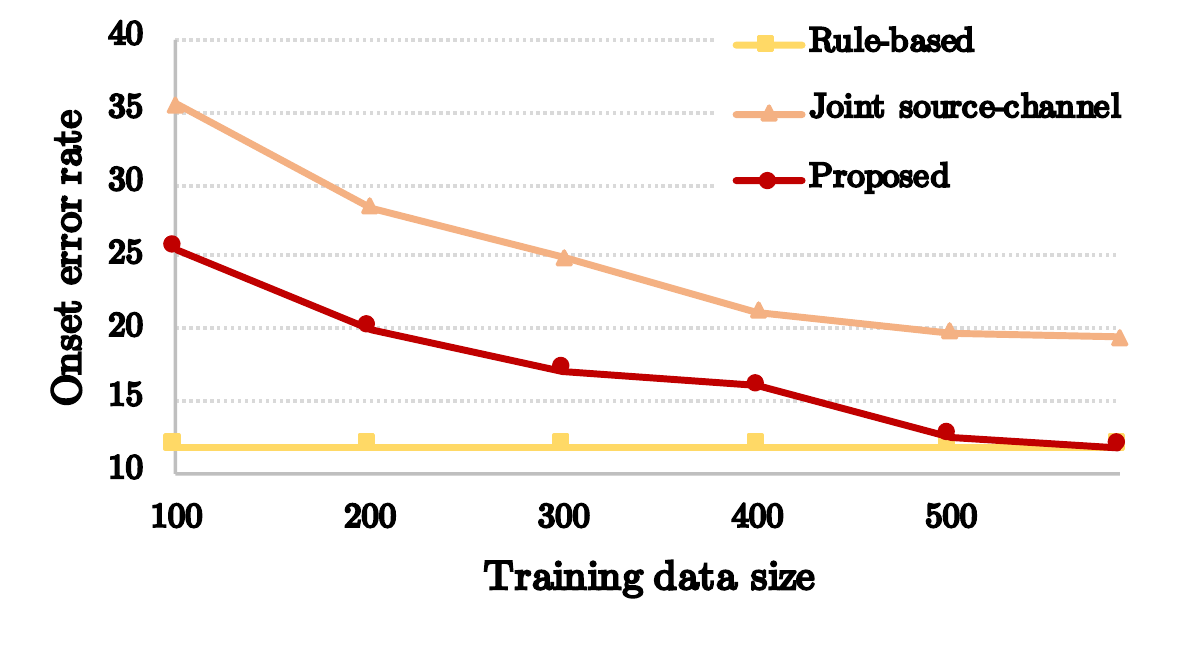}}
    \label{fig:kws_onset}
  \end{subfigure}
  \begin{subfigure}[t]{0.3\textwidth}
    \centering
    \caption{Nucleus error rate}
    \vspace{-0.2cm}
    \centerline{\includegraphics[height=4.2cm]{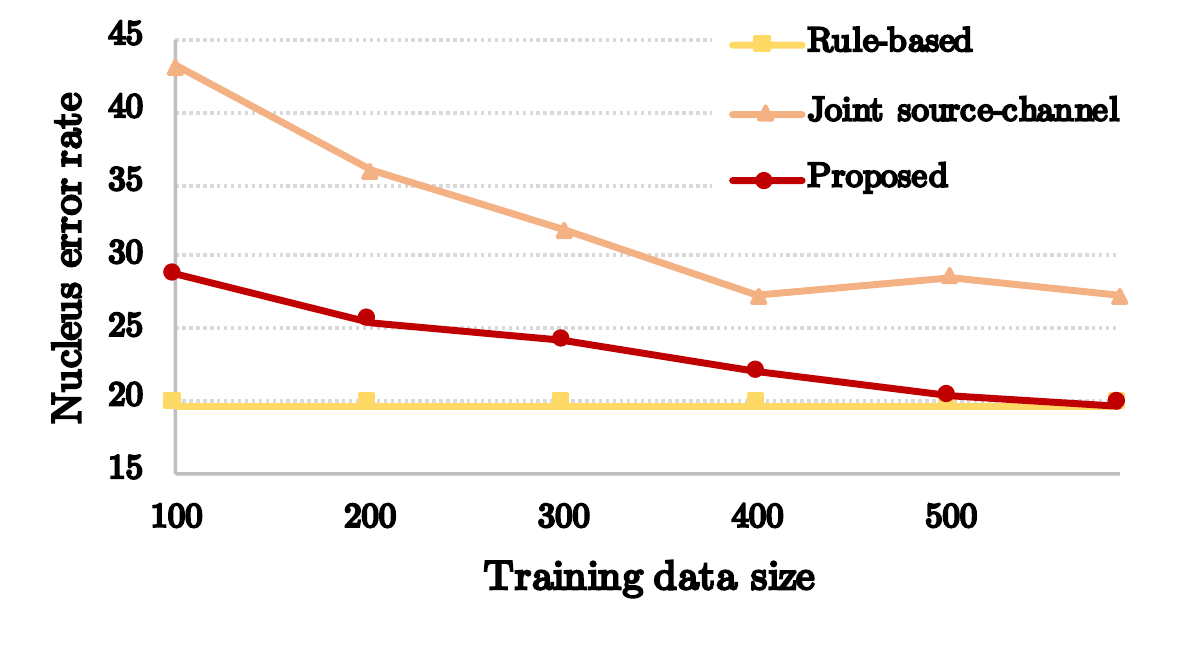}}
    \label{fig:kws_nucleus}
  \end{subfigure}
  \begin{subfigure}[t]{0.3\textwidth}
    \centering
    \caption{Coda error rate}
    \vspace{-0.2cm}
    \centerline{\includegraphics[height=4.2cm]{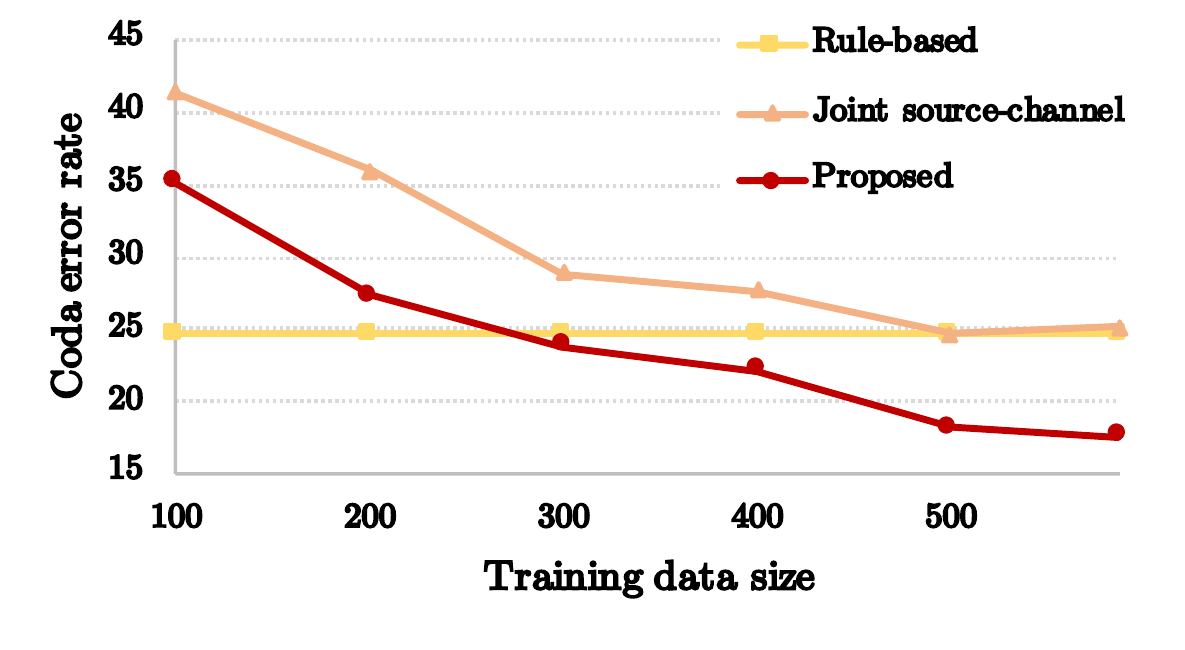}}
    \label{fig:kws_coda}
  \end{subfigure}
  \begin{subfigure}[t]{0.3\textwidth}
    \centering
    \caption{Tone error rate}
    \vspace{-0.2cm}
    \centerline{\includegraphics[height=4.2cm]{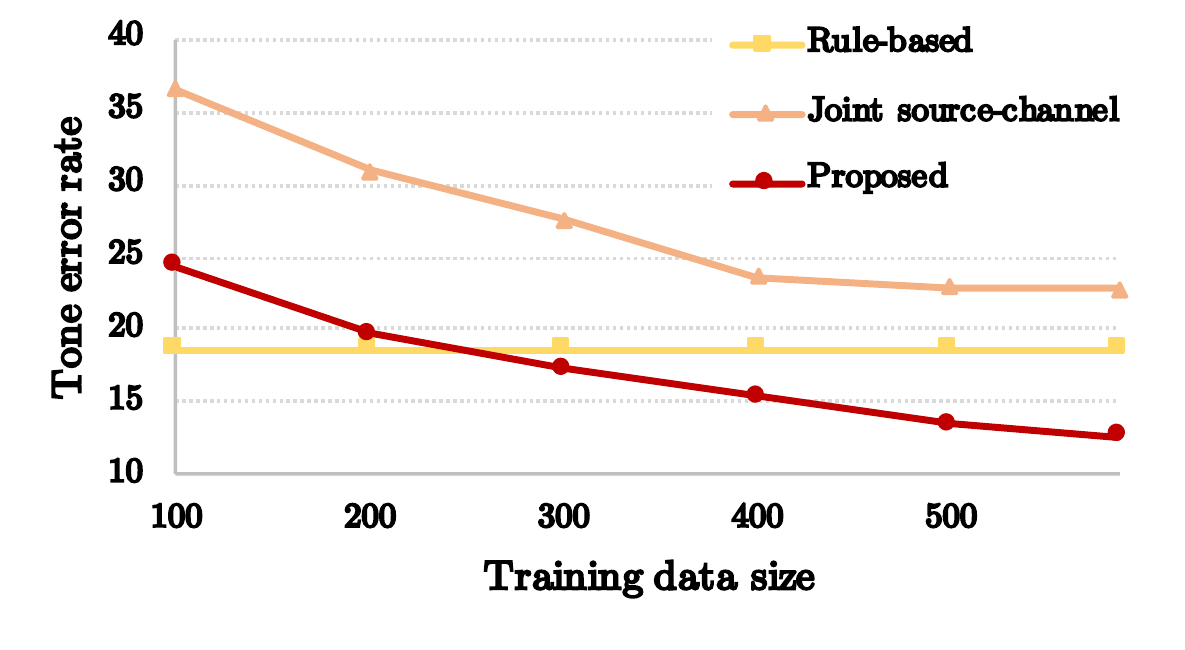}}
    \label{fig:kws_tone}
  \end{subfigure}
  \caption{Performance of different transliteration models as a function of corpus size (Vietnamese dataset - NIST OpenKWS13 dataset).}    
  \label{fig:vie_detailed_performance}
\end{figure}

\begin{figure}[ht]
  \centering
    \begin{subfigure}[b]{0.3\textwidth}
    \centering
    \caption{Syllable error rate}
    \vspace{-0.2cm}
    \centerline{\includegraphics[height=4.2cm]{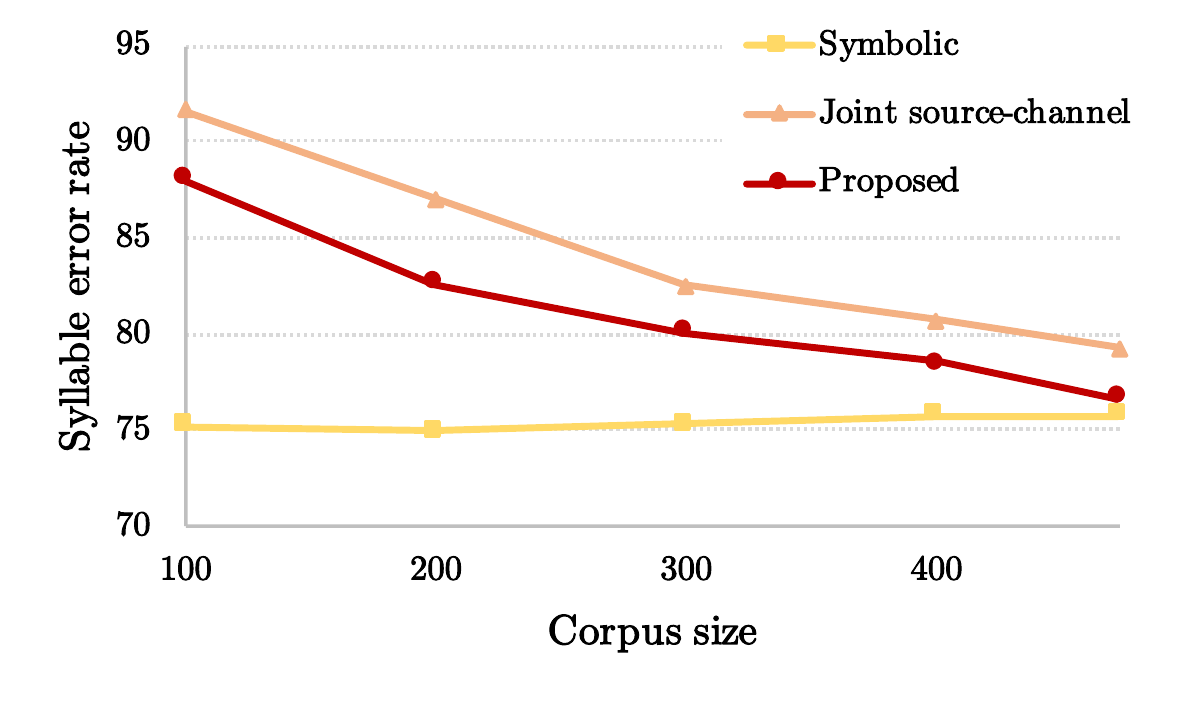}}
    \label{fig:hkpoly_syler}
  \end{subfigure}
  \begin{subfigure}[b]{0.3\textwidth}
    \centering
    \caption{Onset error rate}
    \vspace{-0.2cm}
    \centerline{\includegraphics[height=4.2cm]{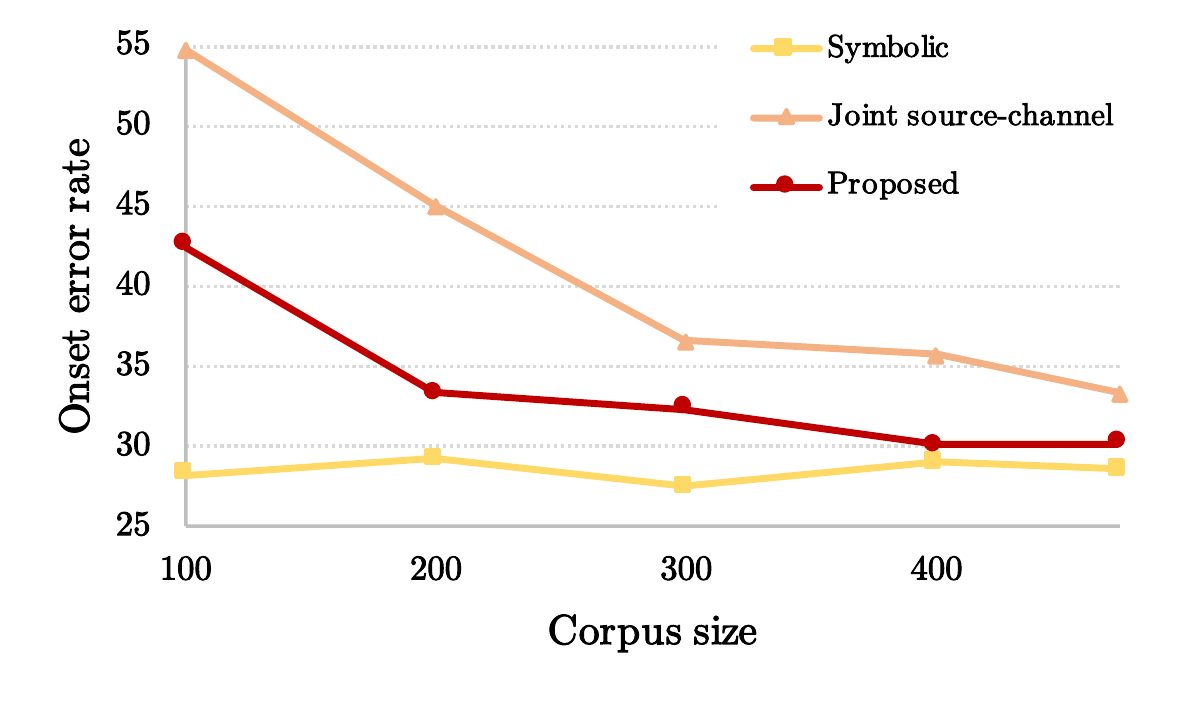}}
    \label{fig:hkpoly_onset}
  \end{subfigure}
  \begin{subfigure}[t]{0.3\textwidth}
    \centering
    \caption{Nucleus error rate}
    \vspace{-0.2cm}
    \centerline{\includegraphics[height=4.2cm]{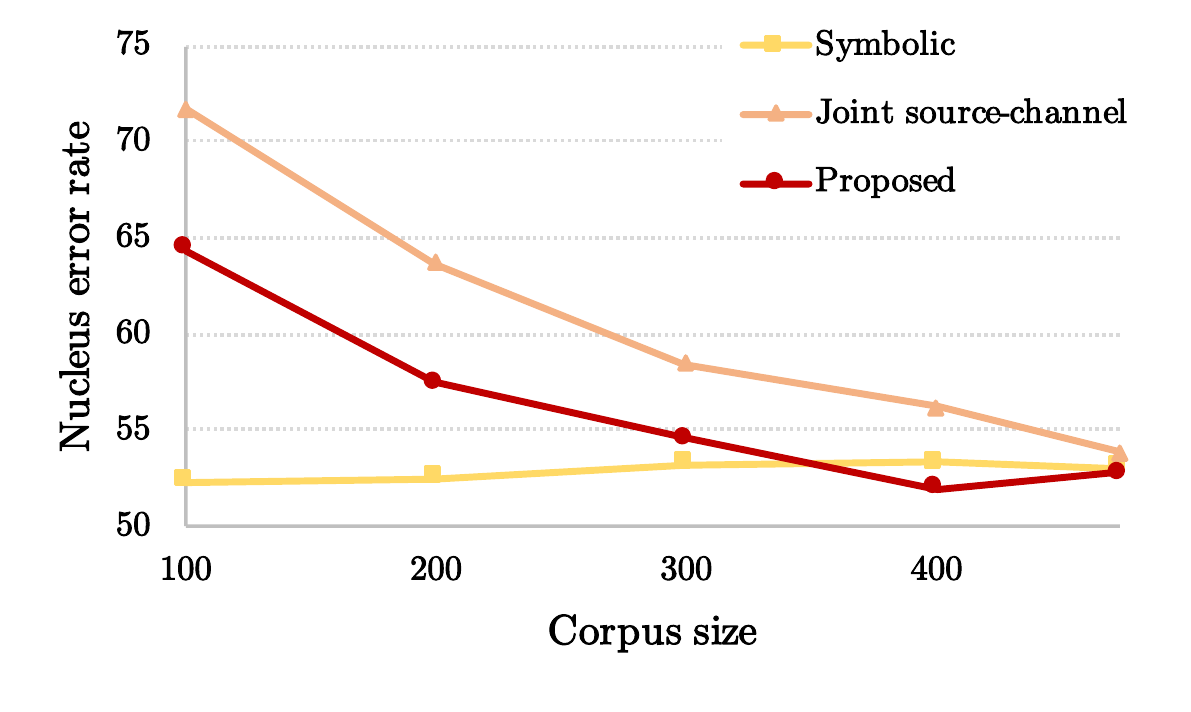}}
    \label{fig:hkpoly_nucleus}
  \end{subfigure}
  \begin{subfigure}[t]{0.3\textwidth}
    \centering
    \caption{Coda error rate}
    \vspace{-0.2cm}
    \centerline{\includegraphics[height=4.2cm]{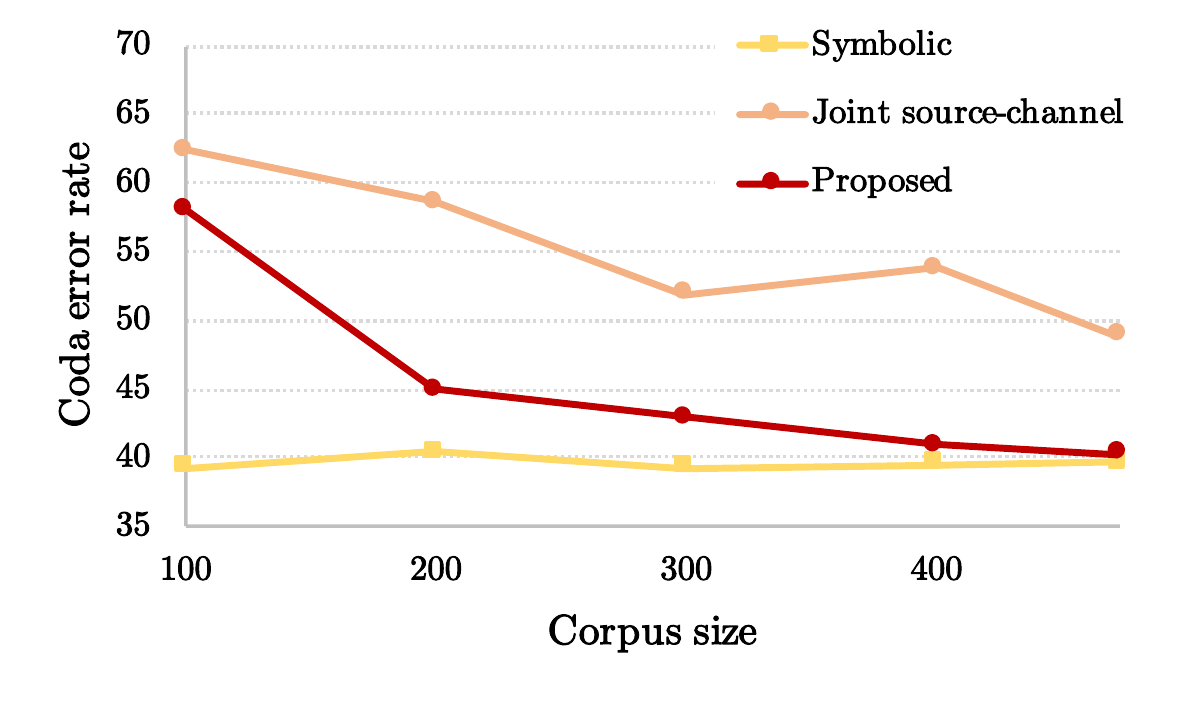}}
    \label{fig:hkpoly_coda}
  \end{subfigure}
  \begin{subfigure}[t]{0.3\textwidth}
    \centering
    \caption{Tone error rate}
    \vspace{-0.2cm}
    \centerline{\includegraphics[height=4.2cm]{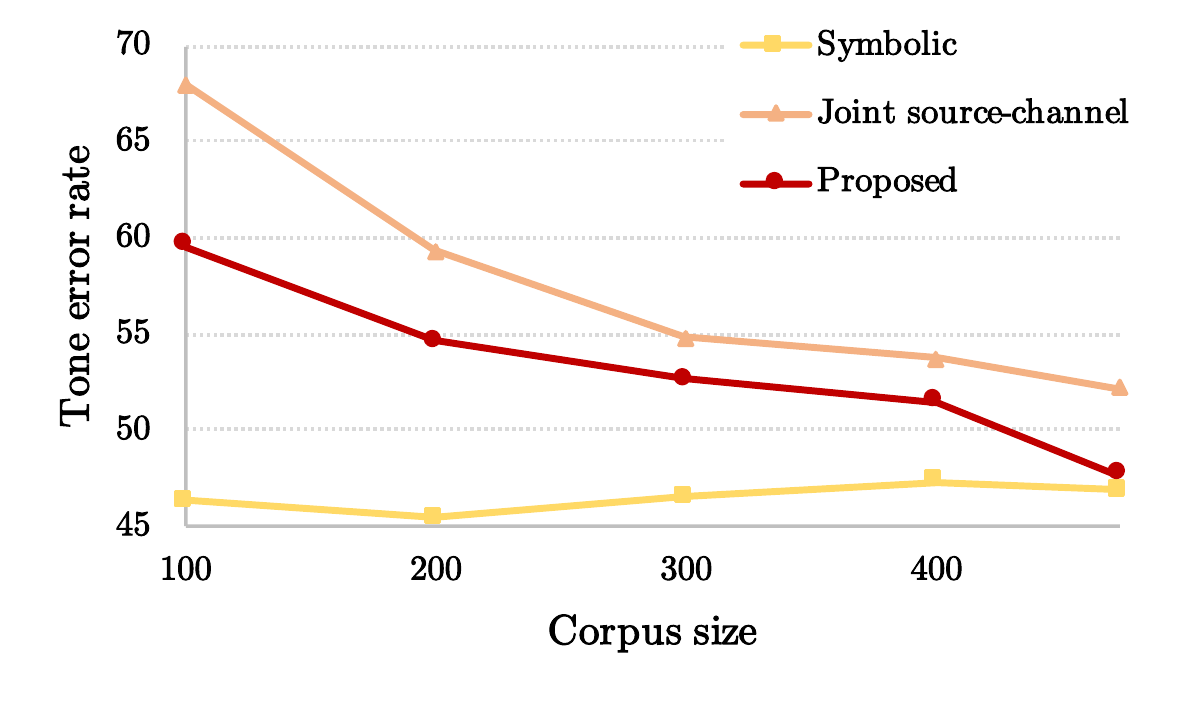}}
    \label{fig:hkpoly_tone}
  \end{subfigure}
  \caption{Performance of different transliteration models as a function of corpus size (Hong Kong Polytechnic Cantonese dataset).}    
  \label{fig:cantonese_detailed_performance}
\end{figure}
  
\subsection{Further Analysis on Syllabic and Sub-Syllabic Structure}
Since the phonological knowledge of syllabic structures are used to better guarantee well-formed syllables in the transliteration output, we want to have a more detailed analysis of improvement by the proposed model over traditional statistical model at individual sub-syllabic positions.

  a) Syllabic error rate: A syllable is the combination of tokens (for both phonemes and tones) within a pair of syllable's delimiters. Each syllable is treated as a token and  the error rate was computed as token error rate using SCLITE.

  b) Onset, Nucleus, Coda, and Tone error rate: To compute the syllable error rate, the hypothesis and reference syllables are aligned to minimize the error distance between the two sequences. From these aligned sequences, output and reference syllables with the same number of sub-syllabic units are extracted. A sub-syllabic unit is counted as incorrect if it is different from the corresponding reference unit. Note that since the error rates of sub-syllabic units are only computed for the subset of output and reference syllables with the same number of units, we only approximate the actual error rates of the sub-syllabic units.

  \subsubsection{Vietnamese}
  \label{subsubsec:vie_further_analysis}
   As shown in Figures \ref{fig:kws_syler} - \ref{fig:kws_tone}, the proposed model consistently outperforms the statistical baseline for all sub-syllabic units across all the training set sizes for Vietnamese.
  At the smallest training data size (100 entries), the proposed model improves the joint source-channel model by 16.62\% relative for syllable error rate, 27.97\% relative for onset, 33.29\% relative for nucleus, 15.25\% relative for coda, and 33.51\% relative for tone error rate.
  
  At the largest training data size (587 entries), the proposed model improves the joint source-channel model by 22.33\% relative for syllable error rate, 39.06\% relative for onset, 27.52\% relative for nucleus, 29.90\% relative for coda, and 44.68\% relative for tone error rate.
The proposed model also improves the symbolic system by 1.42\% relative for syllable error rate, 28.57\% relative for coda error rate, and 32.32\% relative for tone error rate.

  \subsubsection{Cantonese}
  \label{subsubsec:cantonese_further_analysis}
   As shown in Figures \ref{fig:hkpoly_syler} - \ref{fig:hkpoly_tone}, the proposed model consistently outperforms the statistical baseline for all sub-syllabic units across all the training sets for Vietnamese.
  At the smallest training data size (100 entries), the proposed model improves the joint source-channel model by 3.92\% relative for syllable error rate, 22.37\% relative for onset, 10.27\% relative for nucleus, 6.84\% relative for coda, and 12.34\% relative for tone error rate.
  
  At the largest training data size (473 entries), the proposed model improves the joint source-channel model by 3.33\% relative for syllable error rate, 9.38\% relative for onset, 2.06\% relative for nucleus, 17.45\% relative for coda, and 8.69\% relative for tone error rate.

\section{Discussion}

\subsection{Statistical modeling with phonological knowledge}
The proposed framework 
can be interpreted as a statistically grounded framework that adopts symbolic transliteration concepts.
The proposed framework uses pseudo-syllables to define the general constraints on the structure of the transliteration output: how graphemes of the source word should conform to the phonology of the target language's syllables.
From the training data, the framework learns specific distributions of the combinations of graphemes that constitute the pseudo-syllables,
the mapping of individual units of the pseudo-syllable to the target language's phonemes,
and the assignment of tones to each of the syllables.
As a result, the transliteration performance of the proposed framework improves as the data sizes increase while the performance of the symbolic model is constant across all data sizes, as shown in Figure \ref{fig:vie_performance} - \ref{fig:cantonese_detailed_performance}.

In \cite{ngo2015phonology}, we offered a more detailed analysis on the trade-off between data size and performance for the different models.
Using a larger corpus (HCMUS corpus \cite{cao2010}), we compared the transliteration performance using TER and SER for the proposed framework, the standard joint-source channel model and the symbolic model.
The proposed framework performs better than the symbolic system in TER for all corpus sizes larger than 350 words, and better than the symbolic system in all three metrics for all corpus sizes larger than 1,000 words.

While symbolic frameworks for transliteration are still highly valuable when little training data is available, they are also non-trivial to derive.
We constructed two symbolic frameworks for Vietnamese and Cantonese with the idea of sharing as much general basic principles between them as possible.
Nonetheless, while we tried our best to optimize their performance in terms of the string error rate (Figure \ref{fig:kws_ser} and Figure \ref{fig:cantonese_ser}), there are obviously many cases that our rules failed to capture, as evident by the high token error rate (Figure \ref{fig:kws_ter} and Figure \ref{fig:cantonese_ter}) and sub-syllabic unit error rates (Figure \ref{fig:kws_syler} - Figure \ref{fig:kws_tone} and Figure \ref{fig:hkpoly_syler} - Figure \ref{fig:hkpoly_tone}) as compared to the statistical models.
Such difficulties would definitely render it highly challenging to extend a symbolic transliteration framework to other languages, or even different dialects of the same language.

The proposed approach ensures the transliteration output is valid, as specified by the target language's phonology. 
Figure \ref{fig:kws_syler} and \ref{fig:hkpoly_syler} show that the proposed framework consistently improves the syllables' error rate of the baseline statistical model across all languages and all data sizes.
This evidence verifies the strength of the proposed approach.
By augmenting a statistical model for transliteration with phonological knowledge, the proposed model better captures the syllabic structures and syllabic boundaries in scenarios with limited linguistic resources.

\subsection{Future Extensions}

In our analysis of data size vs. performance trade-off between different transliteration models for Vietnamese language in \cite{ngo2015phonology}, the phonology-augmented statistical framework performs consistently better than the joint source-channel model for corpora of size up to 1,500 word pairs.
The  joint source-channel model caught up with the proposed framework for corpora of size 1,500 word pairs and more.
In our analysis for other languages with readily available large datasets for transliteration such as Mandarin and Japanese, we observe a similar trend.
We suspect that such performance differences could be due to different model assumptions. 
Below we elaborate on some future extensions to potentially improve the proposed framework. 

\subsubsection{Modeling inter-syllable context} 

In the proposed framework, pseudosyllable-to-phoneme mapping and tone assignment is contained within each syllable's boundary.
From our preliminary analysis, it is possible that inter-syllable context of phonemes may play some role in determining grapheme to phoneme mapping, as well as how a tone can be assigned to a syllable. Without considering syllable's boundaries, the traditional transliteration model can capture such inter-syllabic context more easily and produces better performance given sufficient training data.
 
\subsubsection{Beyond monosyllabic tonal languages}
In this work, we focus on monosyllabic tonal languages that share a similar phonological structure and lexical tones. The proposed model can be applied to languages such as Korean which, while not tonal, share similar syllable structures as Vietnamese and Cantonese. 

\subsubsection{End-to-end modeling}
Each of the three steps of the proposed model is optimized separately, which means an error in one step is propagated to the next with no mechanism to correct the mistake in a later step.
The statistical baseline models syllabic segmentation, grapheme to phoneme and tone assignment together using the joint source-channel sequences, and thus, performs optimization for all the steps together. Approaches that we are currently exploring include generative models such as Bayesian graphical models, variational autoencoders, and generative adversarial networks (GANs). 

\section{Conclusion}
We proposed a phonology-augmented statistical framework for transliteration.
Using phonological knowledge to augment the statistical n-gram language modeling, our proposed framework ensures the transliteration outputs are valid, resulting in lower error rates compared with baseline approaches in limited-resources scenarios.
On the other hand, using the concept of pseudo-syllables as an additional phonological constraint, while being largely generic and language-independent, offers an approach to automate the learning of syllable formulation instead of relying on exhaustive effort to build predefined symbolic systems.

\section*{Acknowledgment}
The authors would like to thank Mr. Risheng Gao's expertise in Cantonese and Mandarin in helping us preprocess the data. The authors would also like to thank Dan Jurafsky and Mark Hasegawa-Johnson, Haizhou Li, and Bin Ma for the insightful discussions that helped improve the manuscript. 
\appendices
\section{Symbolic Framework}
\label{app:rule_based}
Our symbolic cross-lingual framework was first proposed for Vietnamese language in \cite{ngo2014minimal}.
The proposed framework was mainly a phoneme-to-phoneme system with the source words' phones produced by a text-to-phoneme tool for American English \cite{cmu_t2p}, before being mapped to Vietnamese phonemes.
We have since extended the symbolic framework to Cantonese by directly using the source words' graphemes for the conversion.
Using the source words' graphemes, we can make use of the richer context of the graphemes without making strong assumptions on the source words' origin.
On the other hand, the algorithm of the symbolic framework remains the same.
\subsection{Outline of the Symbolic Framework for Cantonese}
Input: graphemes of a source word.

Output: a sequence of phonetic tokens of the target language that includes phonemes, syllabic delimiters and lexical tones.
Cantonese phonemes are represented in Jyutping \cite{jyutping}.

\begin{enumerate}
  \item{Syllable-splitting:}
Similar to Vietnamese, Cantonese is also a monosyllabic tonal language \cite{gao2000acoustic}.
Hence, the strategy used for forming syllables in Cantonese is also similar to that for Vietnamese \cite{ngo2014minimal}.
  \begin{enumerate}
    \item{Segmentation: }
    Graphemes of the source word are segmented into vowel clusters and consonant clusters.
    \item{Role assignment: }
    \par\noindent * From the sequence of segments in the previous step, vowel clusters are assigned to Nucleus units.
    \par\noindent * Consonant clusters are assigned to Onset and Coda units as followed:
        \par\noindent- If the cluster has one consonant, it is assigned to the Onset role because $<$Consonant-Vowel$>$ structure is more common than $<$Vowel-Consonant$>$ or $<$Vowel$>$ structure \cite{carlisle2001syllable}.
        \par\noindent- If the cluster has two consonants or more, the cluster is split into two parts: (1) the first part is assigned to the Coda of the preceding syllable, (2) the second part is assigned to the Onset of the following syllable
    \item{Post-processing:}
      \par\noindent *  Some Vowel clusters are split up to form the Nucleus of two adjacent syllables of the structures $<$Onset-Nucleus$>$, $<$Nucleus-Coda$>$.
  \end{enumerate}

\item Grapheme-to-phoneme mapping:
    Each of the source word's sub-syllabic units are mapped to a target pronunciation's phoneme \cite{silverman1992}\cite{yip1993cantonese}\cite{wan1998automatic}\cite{guo1999mandarin}\cite{yip2002perceptual}\cite{yip2006symbiosis}.
There are also more specific mapping rules that use the context of preceding and following sub-syllabic units of the source word's graphemes to perform the mapping.

\item Lexical tone assignment:
A lexical tone is assigned to each of the target pronunciation's syllables based on its phones.
\end{enumerate}

\begin{figure}[!t]
\vspace{-0.5cm}
\centering
\includegraphics[width=\linewidth]{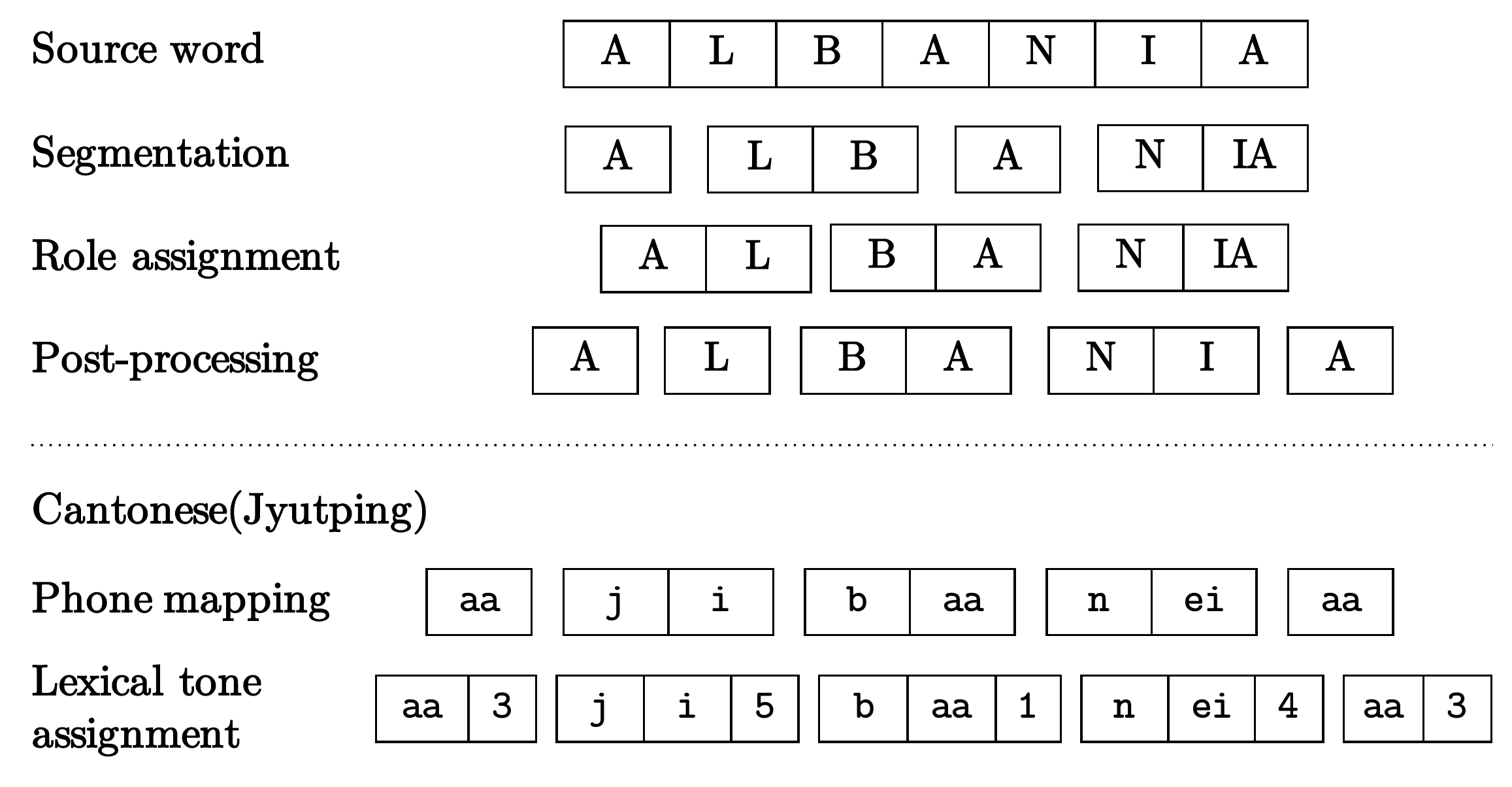}
\caption{Example of the symbolic transliteration framework for Cantonese}
\vspace{-0.5cm}
\label{fig:rule_based_example}
\end{figure}

Figure \ref{fig:rule_based_example} shows how the world \textit{ALBANIA} is transliterated to Cantonese under the symbolic framework.
\subsection{Compensation Strategies}
When converting words from one language to pronunciation in another, there are compensations need to be made such that the output pronunciation does not only best retain the acoustic authenticity of the source word, but also conforms to the target language's phonology.
We present in the following sections some of the compensation strategies used for Cantonese transliteration, with respect to each sub-syllabic unit.
\subsubsection{Onset}
Since Cantonese does not have consonant clusters, these clusters are split, with the epenthesis vowels added to form new syllables\cite{yip1993cantonese}\cite{guo1999mandarin}.
\\\\
\begin{footnotesize}
\begin{tabular}{| p{2.5cm} | p{5.6cm} |}
\hline
English  & \textbf{gr}eenland  \\ \hline
Cantonese (Jyutping) & \verb|g aa k 3 . l i ng 4 . l . aa . n 4| \\ \hline
\end{tabular}
\end{footnotesize}
\subsubsection{Coda}
For consonant clusters in Coda role, both epenthesis and deletion are valid compensation strategies for Cantonese language \cite{guo1999mandarin}\cite{wee2006syllabification}.
For example, given the source word \textit{BOLT}, vowel insertion occurs for both \textbf{l} and \textbf{t}.
\\\\
\begin{footnotesize}
\begin{tabular}{| p{2.5cm} | p{5.6cm} |}
\hline
English & bo\textbf{lt} \\ \hline
Cantonese (Jyutping) & \verb|b o 1 . j i 5 . d a k 6| \\ \hline
\end{tabular}
\end{footnotesize}
\\
However, given the source word \textit{FORD}, vowel insertion only applies to \textbf{d} while \textbf{r} is deleted.
\\\\
\begin{footnotesize}
\begin{tabular}{| p{2.5cm} | p{5.6cm} |}
\hline
English & fo\textbf{rd} \\ \hline
Cantonese (Jyutping) & \verb|f u k 1 . d a k 6| \\ \hline
\end{tabular}
\end{footnotesize}
\\
For consonant clusters where the first letter is a liquid, for example \textbf{r} in the consonant cluster \textbf{rt}, the liquid is usually deleted.
This is because liquid is not salient compared to its neighboring phones and therefore is not perceived by Cantonese speakers \cite{silverman1992}\cite{yip1993cantonese}\cite{yip2002perceptual}.
Thus, the liquid is likely to be deleted in the output pronunciation.
Furthermore, in Cantonese, Codas are either stop or nasal \cite{silverman1992}.

\subsection{Lexical tones}
There are some patterns for lexical tones observed among the Cantonese loanwords in our experimental data.
\begin{itemize}
\item Syllables with ``p'', ``t'' or ``k'' as Coda only accept tone 1, 3 or 6.
\item 95\% of syllables with ``p'' as Coda have either tone 3 or tone 6.
\item 90\% of syllables with ``m'' as Coda have either tone 1 or tone 4.
\end{itemize}




\bibliographystyle{IEEEtran}
\bibliography{references}

\begin{IEEEbiography}
[{\includegraphics[width=1in,height=1.25in,clip,keepaspectratio]{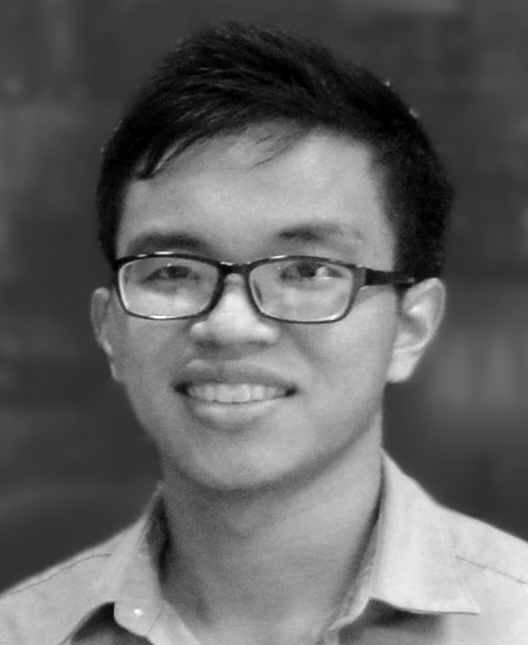}}]{Gia H. Ngo}
received his B.Eng from the National University of Singapore (NUS) in 2015 and currently pursuing his PhD at Cornell University.
Gia has worked at Human Language Technology department at the Institute of Infocomm Research, A*STAR (2013), the Computational Brain Imaging Group at National University of Singapore (2016 - 2017), and GIVE.asia (2014 - 2018). 
Gia's research interests lie in machine learning, Bayesian statistics and their application in natural language processing and computational neuroscience.
In particular, Gia's research interests in natural language processing include transliteration in limited resource scenarios and Bayesian graphical model in language modeling.
His research interests in computational neuroscience include parametric mixture models and non-parametric hierarchical Bayesian models to estimate reference atlases of brain networks from large datasets of neuroimages.

\end{IEEEbiography}
\begin{IEEEbiography}
[{\includegraphics[width=1in,height=1.25in,clip,keepaspectratio]{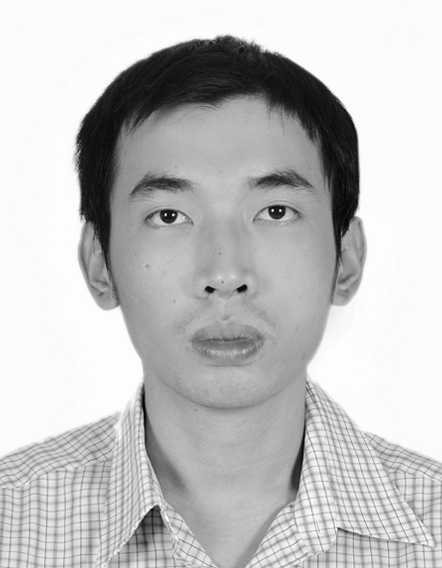}}] {Minh Nguyen}
received his B.Eng from the National University of Singapore (NUS) in 2016.
He is currently a research assistant at the Institute for Infocomm Research (I2R),
Singapore and Clinical Imaging Research Center (CIRC), Singapore.

Minh's research interests include the application of machine learning to problems in natural language processing, computational neuroscience and robotics. For natural language processing, Minh is interested in using machine learning, including deep learning techniques, to model the phonology of loanwords in different languages.
Minh's research interests in computational neuroscience include applying time series modeling to predict diseases progression in human neural system
and to denoise neuroimaging data.

\end{IEEEbiography}
\begin{IEEEbiography}
[{\includegraphics[width=1in,height=1.25in,clip,keepaspectratio]{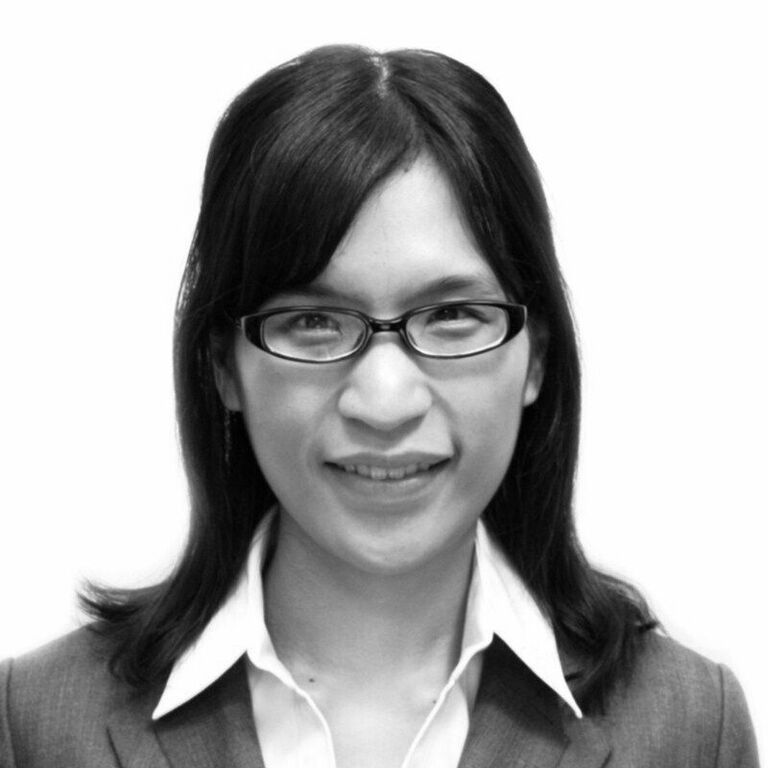}}]{Nancy  F.  Chen  (S'03-M'12-SM'15)} received her Ph.D. from MIT and Harvard in 2011. She worked at MIT Lincoln Laboratory on her Ph.D. research in multilingual speech processing. She is currently leading initiatives in deep learning, conversational AI, human language technology, and Cognitive Human-Like Empathetic and Explainable Machine Learning (CHEEM) at I2R, A*STAR and A*AI, Singapore. She is an adjunct faculty member at Singapore University of Technology and Design. Dr. Chen led a cross-continent team for low-resource spoken language processing, which was one of the top performers in the NIST Open Keyword Search Evaluations (2013-2016), funded by the IARPA Babel program. Dr. Chen is an elected member of IEEE Speech and Language Technical Committee (2016-2018) and was the guest editor for the special issue of ``End-to-End Speech and Language Processing'' in the IEEE Journal of Selected Topics in Signal Processing (2017). Dr. Chen has received multiple awards, including Best Paper at APSIPA ASC (2016), the Singapore MOE Outstanding Mentor Award (2012), the Microsoft-sponsored IEEE Spoken Language Processing Grant (2011), and the NIH Ruth L. Kirschstein National Research Award (2004-2008). 
\end{IEEEbiography} 
\end{document}